**Title:** A fully articulated humanoid dexterous hand with human-level joint structure enabled by linkage-based actuation

**Authors:**
Jihao Li[1,2], Jianguo Wang[1,2], Haotian Guo[1], Yanzhe Wang[1], Guodong Lu[2], Huixu Dong[1,2]*

**Affiliations:**
[1]Grasp Lab, School of Mechanical Engineering, Zhejiang University, Hangzhou, China.
[2]Robotics Institute of Zhejiang University, Ningbo, China.
*Corresponding author. Email: huixudong@zju.edu.cn

**Abstract:** Humanoid dexterous hands are a core component of embodied intelligence, enabling robots to translate high-level intent into rich and adaptive physical interaction. However, existing designs remain constrained by a persistent trade-off among anthropomorphic kinematic complexity, mechanical robustness, load capacity and control simplicity, limiting dexterous manipulation in unstructured environments. Here we present Torch-Hand, a fully articulated humanoid dexterous hand that replicates the joint structure and motion hierarchy of the human hand with 21 degrees of freedom (DOF), comprising a highly articulated five-DOF thumb and four DOF in each of the remaining fingers. The differentiated thumb integrates multi-axis CMC motion to provide broad opposition and flexible contact orientation relative to the four fingers. Unlike tendon-driven or hybrid actuation approaches, Torch-Hand adopts an entirely linkage-based transmission architecture. By integrating screw-linkage actuation, the hand achieves high grasping force and substantial payload capacity while maintaining a compact anthropomorphic form. Its actuation architecture is optimized for control tractability, with all actuators integrated within the palm and arranged in a serial-robot-like configuration that establishes a direct correspondence between motor motion and joint kinematics. This architecture mechanically decouples functional DOF to reduce unintended inter-joint motion propagation, while allowing independently addressable DOF to be actively coordinated into coupled multi-joint motion patterns during complex manipulation. Comprehensive experiments assess reachable workspace, coordinated fingertip trajectories, thumb opposition, 33-grasp taxonomy coverage, grasping force, payload capacity and task-level manipulation. These results establish Torch-Hand as a robust hardware foundation for dexterous embodied manipulation.

**One-Sentence Summary:** Torch-Hand unites decoupled dexterity, linkage-driven strength, and coordinated control for humanoid manipulation.

**Main Text:**

## INTRODUCTION

Embodied intelligence has emerged as a central research direction in robotics, accelerating the development of humanoid systems capable of physical interaction in contact-rich, unstructured settings (1-5). In these systems, the dexterous hand serves as one of the most critical interfaces between robot and world (6), converting intent into precise, adaptive action. Practical manipulation performance requires more than anthropomorphic morphology or a higher number of joints. Rather, a compact humanoid hand must support both delicate fingertip manipulation (7,8) and powerful whole-hand grasps (9,10), combining high motion precision with substantial load-bearing capacity. The human hand provides the natural benchmark for such integration, as these otherwise competing capabilities coexist within a compact anatomical system. Such integration is particularly indispensable in precision assembly under load, where substantial forces must be sustained without compromising fine motion control. For instance, mounting a heavy camera lens onto a threaded optical mount requires the hand to support the lens's weight, precisely align the optical axis, and dexterously rotate the lens into place.

The pursuit of human-like dexterity, motion precision and load-bearing capacity has motivated diverse structural strategies in humanoid dexterous hand design, broadly aimed at reproducing or surpassing key functions of the human hand in applications spanning healthcare, industrial manufacturing and human-robot interaction (11-13). Rigid motor-gear hands, exemplified by the DLR-HIT hand, provide precise and deterministic finger motion (14), but embedded motors and gears increase bulk, complicating high-DOF integration for anthropomorphic dexterity, while thermal constraints limit sustained load-bearing capacity. Pneumatic soft hands exploit compliance for passive adaptability and interaction safety, but coupled deformation and low structural stiffness limit controllable dexterity, motion precision and sustained force output (15). Tendon-driven rigid-soft hybrid hands instead provide a more scalable route to anthropomorphic dexterity through compact multi-DOF integration. Zhang et al. employed biomimetic skeleton-ligament structures to achieve controllable multi-DOF mobility with structural and control simplicity (16), whereas Chen et al. relocated actuation to the forearm to realize 18-DOF whole-hand mobility (17). Nevertheless, friction and transmission efficiency continue to limit motion precision and load-bearing capacity. Rather than directly reproducing biological musculoskeletal architectures, linkage-based systems offer a promising engineering-oriented route to reconciling dexterity, motion precision and load-bearing capacity. The HIT-VI hand employed a miniature compound coupled-adaptive linkage mechanism to enable adaptive grasping, with mechanical self-locking further enhancing load-bearing capacity and grasp stability (19). Using parallel-mechanism-inspired spatial linkages, the ILDA hand demonstrated dexterous manipulation with 15 active DOF while delivering fingertip forces of up to 34 N per finger (18).

Despite recent advances, humanoid dexterous hand design remains constrained by a dexterity-precision-payload trilemma, with no existing compact anthropomorphic architecture yet reconciling all three attributes. This limitation reflects architecture-level compromises inherent to current actuation and transmission strategies. Motor-gear hands offer deterministic joint control and scalable DOF, but embedding motors and reducers within human-scale fingers concentrates distal mass, backlash and packaging density, limiting force density and dynamic robustness (14, 20-30). Pneumatic hands provide distributed compliance and passive adaptability, yet compressibility, nonlinear deformation and low structural stiffness constrain joint-level precision, motion bandwidth and sustained load-bearing (15, 31-40). Tendon-driven hands reduce distal inertia and better approximate biological muscle-tendon routing, but frictional routing, tendon elasticity, slack and fatigue degrade transmission transparency and repeatable force-position control under load (16, 17, 41-49). Hybrid hands combine multiple actuation principles to expand function, but added interfaces introduce compliance, calibration drift and failure modes that undermine robustness (19, 50-58).

Among existing paradigms, linkage-driven architectures offer a particularly promising route toward resolving this trilemma. Rigid linkages provide deterministic kinematic transmission, high structural stiffness and efficient force propagation, enabling multi-DOF anthropomorphic motion without compliant tendons, distal joint motors or low-stiffness fluidic actuators (59-61). However, this potential remains only partially realized. Existing multi-link designs often introduce strongly coupled kinematics (62, 63), depart from human-hand dexterity and morphology (64-66), or remain difficult to integrate into compact hand architectures (67-69). The ILDA hand represents an important advance over previous linkage-based designs (18). Nevertheless, its fixed thumb restricts opposition-based dexterity, and its parallel-robot-inspired architecture increases the complexity of controlling functional DOF.

Motivated by these gaps, we present Torch-Hand (Fig. 1), a fully articulated humanoid dexterous hand that overcomes the dexterity-precision-payload trilemma through a compact, fully linkage-driven anthropomorphic architecture. Supplementary Table S1 summarizes the key structural parameters and performance metrics of Torch-Hand in comparison with representative dexterous robotic hands. The main contributions are summarized as follows:

• High dexterity and compactness. Torch-Hand delivers human-level dexterity in a compact anthropomorphic form factor via a palm-integrated kinematic architecture with 21 DOF. It features a five-DOF opposable thumb and four four-DOF fingers with biomimetic joint kinematics that replicate the functional motion hierarchy of the human hand. All actuators are integrated within the palm to maintain lightweight finger modules and low distal inertia, laying the kinematic foundation for broad thumb opposition, agile fingertip motion, and diverse grasp configurations.

• High precision and decoupled control. Torch-Hand achieves high-precision decoupled control via a control-oriented actuation architecture that separates joint-level mechanical decoupling from motion-level coordination, with a repeatable fingertip positioning accuracy below 0.03 mm. Each functional DOF is mechanically decoupled and independently addressable, establishing a clear actuator-joint-fingertip mapping that improves motion transparency and suppresses unintended inter-joint motion propagation. Critically, this design retains the ability to actively coordinate independent DOFs into coupled multi-joint motions, enabling complex fingertip trajectories, whole-hand grasp postures, and task-level manipulation without sacrificing joint-level controllability. This architecture thus resolves the long-standing trade-off between independent controllability and coordinated motion capability.

• High payload capacity and robustness. Torch-Hand addresses the dexterity-load trade-off through a fully linkage-driven screw-transmission architecture. It delivers industry-leading fingertip load capacity while preserving human-like dexterity across 21 independently actuated DOFs, enabled by a fully linkage-driven screw-transmission architecture. Rigid linkage pathways ensure deterministic force and motion transmission, while screw-driven sliders provide inherent mechanical advantage, positional stability, and self-locking capability. This mechanically constrained design eliminates deformation-induced uncertainty associated with tendon-driven and pneumatic systems, supporting high fingertip force output, strong whole-hand grasping, and substantial payload capacity. The architecture thus unites high-load manipulation, repeatable transmission, and structural robustness within a compact palm-integrated platform.

• System-level capability validation. Torch-Hand is assessed with a validation framework that directly maps onto the three vertices of the dexterity-precision-payload trilemma. Dexterity is evaluated through fingertip workspace mapping, Kapandji test and the 33-grasp taxonomy, probing a capability rarely demonstrated by linkage hands: broad reachability, active thumb opposition and whole-hand grasp diversity in the same mechanism. For precision, trajectory tracking and repeat positioning tests shift the assessment from joint actuation to fingertip-level accuracy, showing that mechanically decoupled DOF and linkage-screw transmission support both repeatable single-DOF positioning and coordinated multi-joint motion. Payload capacity is examined through fingertip force and payload experiments, examining whether high-force transmission can be retained without sacrificing anthropomorphic mobility. Finally, a 3 × 3 task matrix integrates dexterity, precision and payload capacity across fine fingertip operation, object handling and forceful tool use.

The integration of a palm-integrated kinematic layout, mechanically decoupled spatial mechanisms and rigid linkage-screw transmission enables Torch-Hand to reconcile dexterity, precision and payload capacity within a compact anthropomorphic platform. Beyond reconciling three competing performance objectives, this work establishes a compact, fully integrated design paradigm for anthropomorphic dexterous hands, expanding the design space for dexterous manipulation research and enabling new possibilities in precision industrial operation, adaptive prosthetics, and close-proximity human-robot collaboration.

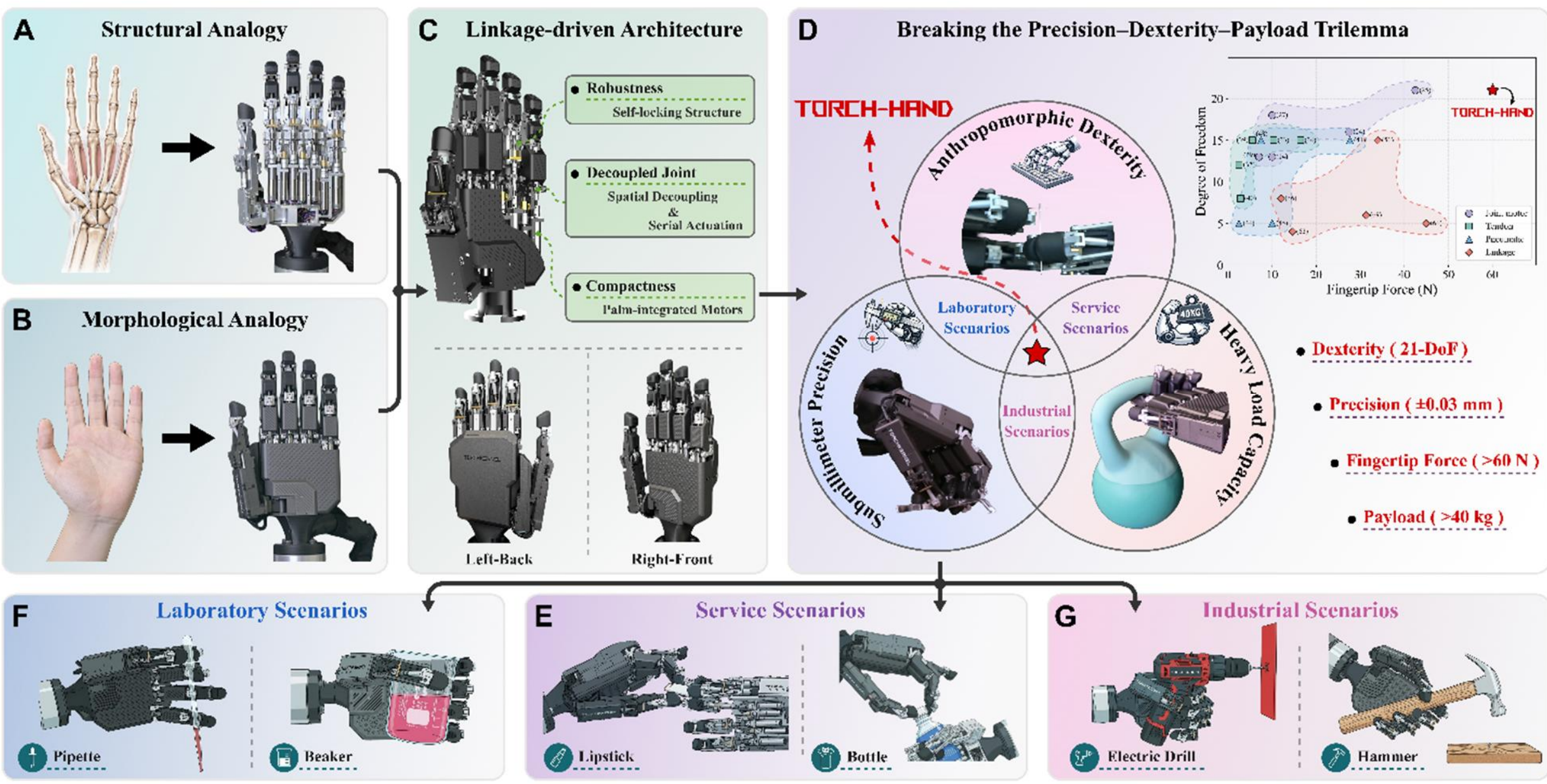

**Fig. 1. Torch-Hand: a fully linkage-driven humanoid hand for dexterous embodied manipulation.** (A and B) Structural and morphological correspondence between the human hand and Torch-Hand. (C) Core technologies and performance advantages of Torch-Hand. (D) Dexterity-precision-payload performance demonstration and comparison against representative robotic hands. (E to G) Representative task scenarios in service, laboratory, and industrial manipulation.

## RESULTS

The integration of a palm-integrated kinematic layout, mechanically decoupled spatial mechanisms and rigid linkage–screw transmission enables Torch-Hand to overcome the dexterity–precision–payload trilemma within a anthropomorphic platform. This architecture supports human-hand-like mobility, repeatable fingertip-level precision and high-force load-bearing capability in a single fully linkage-driven system. Beyond resolving these competing performance objectives, Torch-Hand establishes a compact, fully integrated design paradigm for humanoid dexterous hands, expanding the design space for dexterous manipulation and enabling applications in precision laboratory operations, forceful industrial manipulation and service scenarios.

### THEORY MODEL OF THE DEXTEROUS FINGER

Torch-Hand is anatomically organized into two differentiated modules as shown in Fig. 2A: the four fingers (the index, middle, ring and little fingers) and the thumb module. This modular organization follows the functional asymmetry of the human hand, in which the four fingers primarily support coordinated grasping and manipulation, whereas the thumb enables opposition, expands the reachable workspace and enhances task adaptability. Across both modules, the mechanical design follows a common principle: each functional degree of freedom is assigned an independent actuation pathway. Moreover, All joints in Torch-Hand are constructed from elementary kinematic pairs, and the corresponding link and joint definitions are summarized in Table S2.

Each of the four fingers provides 4 DOF: MCP (Ab/Ad), MCP (F/E), PIP (F/E) and DIP (F/E). As illustrated in Fig. 2B, the motion of each DOF is assigned to an individual slider-crank linkage, with the prismatic joint of the slider serving as the active input. Because each slider-crank linkage operates as a single-input-single-output transmission unit, the position and velocity states of all linkages within the corresponding mechanism can be uniquely determined from the displacement and motion of its active slider. In this baseline architecture, each slider-crank unit is kinematically independent of the others, allowing the DOF of each finger to be regulated independently without inter-unit motion coupling. Thus, the basic finger architecture establishes decoupled actuation and transparent motion mapping at the level of individual functional degrees of freedom.

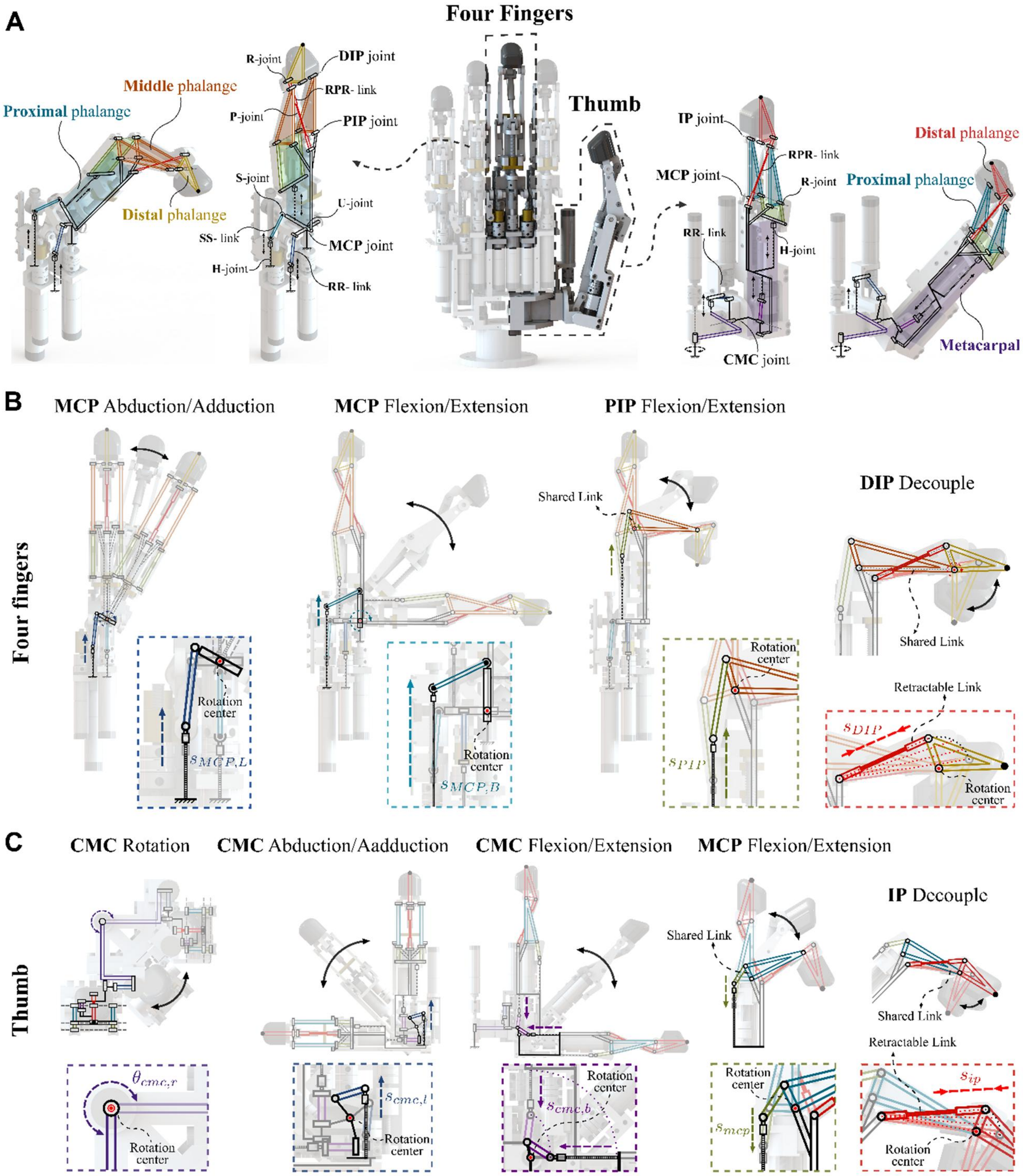

**Fig. 2. Decoupled mechanical architecture of the Torch-Hand.** (A) Overall structural organization of Torch-Hand, showing the differentiated design of the four fingers and the thumb module. (B) Decoupled joint actuation mechanism of the four fingers, including metacarpophalangeal abduction/adduction (MCP Ab/Ad), metacarpophalangeal flexion/extension (MCP F/E), proximal interphalangeal flexion/extension (PIP F/E) and distal interphalangeal flexion/extension (DIP F/E). (C) Decoupled joint actuation mechanism of the thumb, including carpometacarpal rotation (CMC Rot), carpometacarpal abduction/adduction (CMC Ab/Ad), carpometacarpal flexion/extension (CMC F/E), metacarpophalangeal flexion/extension (MCP F/E) and interphalangeal flexion/extension (IP F/E).

Human finger motion, however, is not governed by completely independent joint rotation. In particular, the PIP and DIP joints exhibit pronounced interphalangeal coordination during natural flexion and extension [70-72]. To reproduce this biomechanical feature without compromising controllability, the DIP joint is driven by a four-bar linkage composed of revolute joints rather than by an isolated slider-crank unit. One link of this DIP linkage is shared with the PIP transmission and functions as the input link for DIP motion. As a result, the active slider of the PIP joint can coordinate both PIP and DIP flexion, allowing the finger to reproduce human-like coupled interphalangeal motion during unloaded motion. In this state, the coupling is not an uncontrolled mechanical constraint but a deliberately introduced kinematic relationship that follows the natural coordination of the human PIP and DIP joints.

This PIP-DIP coordination is designed to be condition dependent rather than rigidly fixed. To preserve joint decoupling during grasping and dexterous manipulation, one link in the DIP four-bar mechanism is designed as a compressible link. In the absence of external contact, this link remains at its nominal length and maintains the prescribed coordination between the PIP and DIP joints. When PIP motion is constrained by object contact, or when the DIP joint is independently loaded, the shared link between the PIP and DIP mechanisms remains stationary. Under this condition, the DIP mechanism transitions into an equivalent slider-crank configuration, with the compressible link functioning as the sliding element. This mechanical transition releases the DIP joint from PIP-driven coordination and enables independent DIP motion under contact, thereby combining natural interphalangeal coordination during free motion with task-dependent joint decoupling during object interaction.

The thumb is designed with 5 DOF to preserve its dominant role in grasping and dexterous manipulation: CMC (Rot), CMC (Ab/Ad), CMC (F/E), MCP (F/E) and IP (F/E) (Fig. 2c). The MCP-IP coordination of the thumb follows the same design principle as the PIP-DIP coordination of the four fingers, enabling natural coupled motion during unloaded flexion while allowing contact-dependent decoupling when required. Notably, three rotational DOF are integrated at the thumb CMC joint, with mutually orthogonal motion axes. This arrangement substantially expands the thumb workspace and enhances opposition capability, providing the kinematic basis for human-like grasping, precision manipulation and multi-finger coordination.

Taken together, these designs establish Torch-Hand as an anatomically differentiated hand architecture with independently actuated functional degrees of freedom. This architecture preserves human-like joint coordination during free motion, enables adaptive decoupling during object interaction, and enhances grasp stability, opposition workspace and dexterous manipulation capability through functional differentiation between the four fingers and the thumb.

## KINEMATIC ANALYSIS OF THE DECOUPLED MOTION FINGER

Building on the decoupled mechanical architecture described above, Torch-Hand can be represented as a set of serial kinematic chains extending from the palm to each fingertip. This representation establishes a transparent motion pathway from actuator input to slider displacement, joint rotation and fingertip pose. Unlike tendon-routed or parallel-actuated hands, in which joint motion is often mediated by coupled transmission pathways, each functional degree of freedom in Torch-Hand is assigned a local transmission unit. This design allows joint motion to be described through a direct input-output relationship, providing the kinematic basis for motion transparency in a high-degree-of-freedom hand.

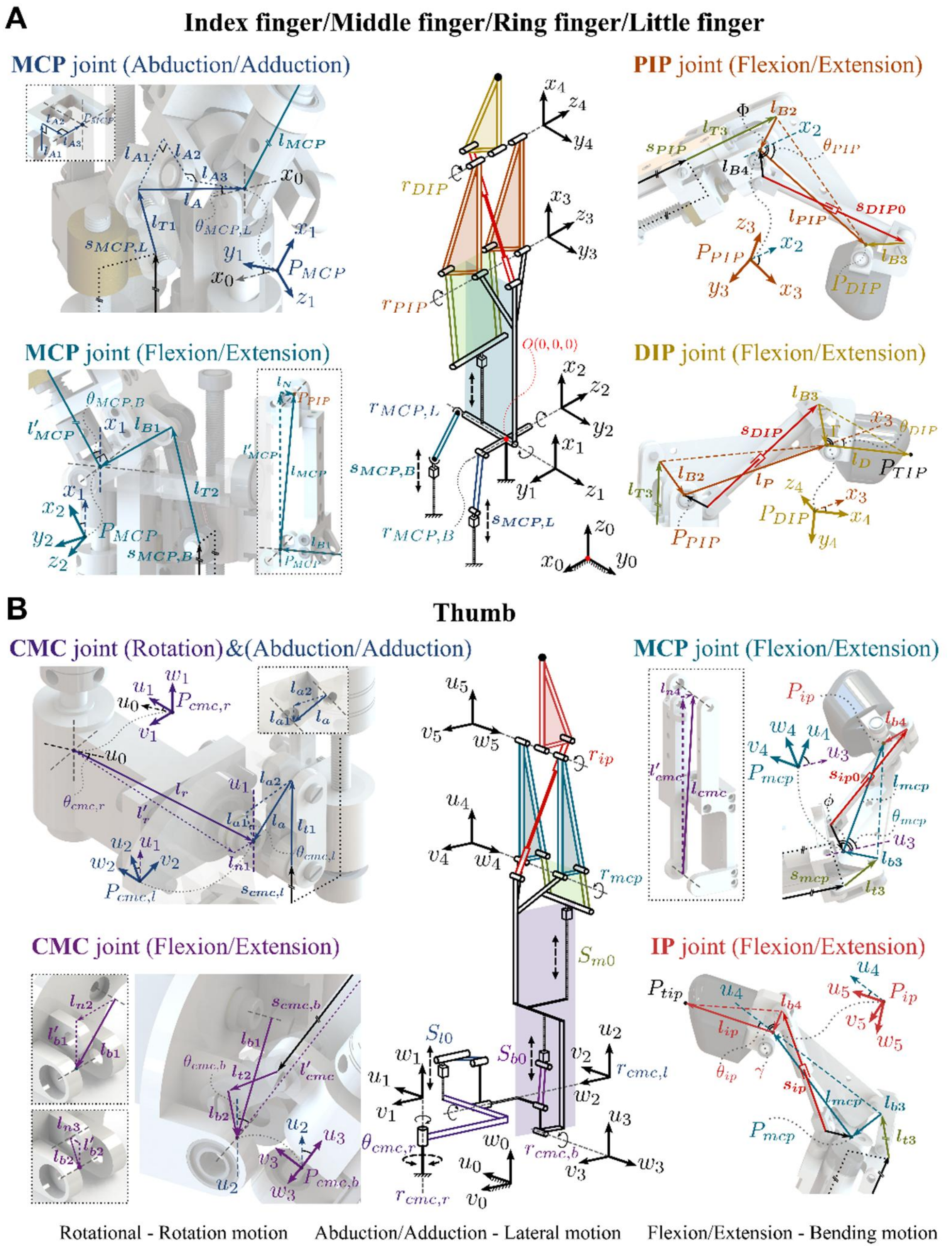

**Fig. 3. Kinematic analysis of the proposed mechanism of Torch-Hand.** (A) Kinematic model of four fingers. (B) Kinematic model of the thumb.

In each local transmission unit, the actuator angular input $\psi_i$ is first converted into the linear displacement $s_i$ of the active slider through the screw-driven transmission. The slider displacement then determines the corresponding joint rotation $\theta_i$ through the geometric closure of the slider-crank linkage, establishing a local mapping from $\psi_i$ to $s_i$ and $\theta_i$. Because this mapping is confined to an individual transmission unit, each actuator-slider-joint pathway remains independently traceable. The resulting joint vector is then introduced into the serial Denavit-Hartenberg (DH) model to compute the fingertip pose. Figure 3 illustrates the serial kinematic models used to describe the four-finger and thumb motions. And the details of the actuator-to-slider mapping $\omega_i \rightarrow s_i$ are provided in Supplementary Eqs. S1-S27 and Supplementary Fig. S1.

For the four fingers part, the joint vector is defined as

$$\mathbf{q}_{\boldsymbol{f}} = [\theta_{MCP,L} \quad \theta_{MCP,B} \quad \theta_{PIP} \quad \theta_{DIP}]^{\mathrm{T}}, \tag{1}$$

while the joint vector of the thumb is

$$\mathbf{q}_{\boldsymbol{t}} = [\theta_{cmc,r} \quad \theta_{cmc,l} \quad \theta_{cmc,b} \quad \theta_{mcp} \quad \theta_{ip}]^{\mathrm{T}}. \tag{2}$$

Here, the subscripts $\boldsymbol{f}$ and $\boldsymbol{t}$ denote the four fingers and the thumb, respectively.For the four fingers, the joint variables are arranged sequentially from MCP (Ab/Ad) and MCP (F/E) to PIP (F/E) and DIP (F/E). For the thumb, the joint variables follow a serial order from CMC (Rot), CMC (Ab/Ad) and CMC (F/E) to MCP (F/E) and IP (F/E). In this notation, $\boldsymbol{L}/\boldsymbol{l}$, $\boldsymbol{B}/\boldsymbol{b}$ and $\boldsymbol{r}$ denote lateral, bending and rotational motions, respectively, following the definitions in the kinematic model.

The serial Denavit-Hartenberg formulation then maps these joint variables to fingertip pose. In the four fingers, each transformation corresponds to one anatomical joint motion along the serial chain. In the thumb, the three mutually orthogonal CMC axes establish a compact serial base for opposition, while the distal MCP and IP joints follow the same local input-output principle used in the four fingers. This architecture expands the thumb workspace without introducing multi-actuator coupling at individual joints. The detailed Denavit-Hartenberg transformations, fingertip pose extraction procedure and complete parameter definitions are provided in Supplementary Eqs. S28-S34 and Supplementary Tables S2 and S3.

The resulting kinematic model preserves motion transparency at both the position and velocity levels. At the joint level, actuator angular input is locally converted into slider displacement and joint rotation without requiring a globally coupled inverse solution. At the fingertip level, the serial-chain representation allows position and differential motion to be computed directly from the anatomical joint variables. Because the actuator-joint mapping remains local and independently defined, changes in one actuator produce predictable motion in the corresponding functional degree of freedom without requiring compensation for unintended motion propagation across separate transmission pathways.

This serial kinematic architecture is particularly important for a high-degree-of-freedom humanoid hand. It preserves the anatomical organization required for human-like motion while avoiding the control burden commonly associated with highly coupled actuation systems. Importantly, decoupled actuation does not impose sequential joint execution. Instead, it provides independent control authority over each functional degree of freedom, allowing multiple joints to be driven simultaneously by their corresponding actuators to reproduce coordinated, fluid and human-like finger movements. Motion transparency therefore arises from the integration of independently addressable joint actuation, serial joint organization and analytically traceable linkage transmission, providing the analytical basis for workspace characterization, fingertip dexterity evaluation and task-level manipulation experiments.

## MECHANICAL STRUCTURE OF TORCH-HAND

The serial and decoupled kinematic architecture of Torch-Hand is physically realized through a fully linkage-driven mechanical system as demonstrated in Fig. 4. This implementation translates the abstract actuator-joint mapping described above into a compact and biomimetic mechanical layout in which each functional degree of freedom is assigned an independent transmission pathway (4A and 4G). The system is organized around three coupled design principles: compact actuator integration, conversion of actuator output into usable joint-driving motion, and mechanical realization of thumb opposition and dexterous functionality.

At the actuator-integration level, all drive units are concentrated within the palm, as illustrated by the palm-side actuator layout in Fig. 4B and 4C. This arrangement preserves the anthropomorphic proportions of the digits and reduces distal inertia, which is essential for fast, stable and precise finger motion. By relocating the drive units away from the phalanges, the fingers remain mechanically lightweight and retain internal space for multi-stage linkage transmissions and potential sensor integration. The palm-concentrated layout also simplifies wiring and electrical routing, reducing integration complexity in a high-degree-of-freedom hand. As a result, Torch-Hand accommodates a large number of independently actuated degrees of freedom without enlarging the distal phalanges or compromising the hand-scale form factor.

This palm-integrated layout requires an efficient mechanism for converting actuator output into precise joint-driving displacement. In Torch-Hand, the rotary output of each actuator is transformed into controlled linear motion of the active slider through a screw-driven transmission (Fig. 4D and 4E). The slider is guided by a dedicated rail-slider pair, which constrains translational motion and improves displacement accuracy. Each revolute joint is further supported by bearings to maintain rotational precision and reduce parasitic friction during repeated motion. The active slider then drives the corresponding linkage unit, converting linear displacement into joint rotation. In addition to providing a compact rotary-to-linear conversion, the screw-driven pathway introduces mechanical gain and self-locking capability. Once the finger reaches a target posture, the self-locking transmission helps sustain external loads without continuous actuator torque, as long as the applied load does not exceed the structural strength of the finger. This combination of guided translation, bearing-supported rotation and screw-based self-locking enables Torch-Hand to achieve precise joint regulation while maintaining high load-bearing capacity.

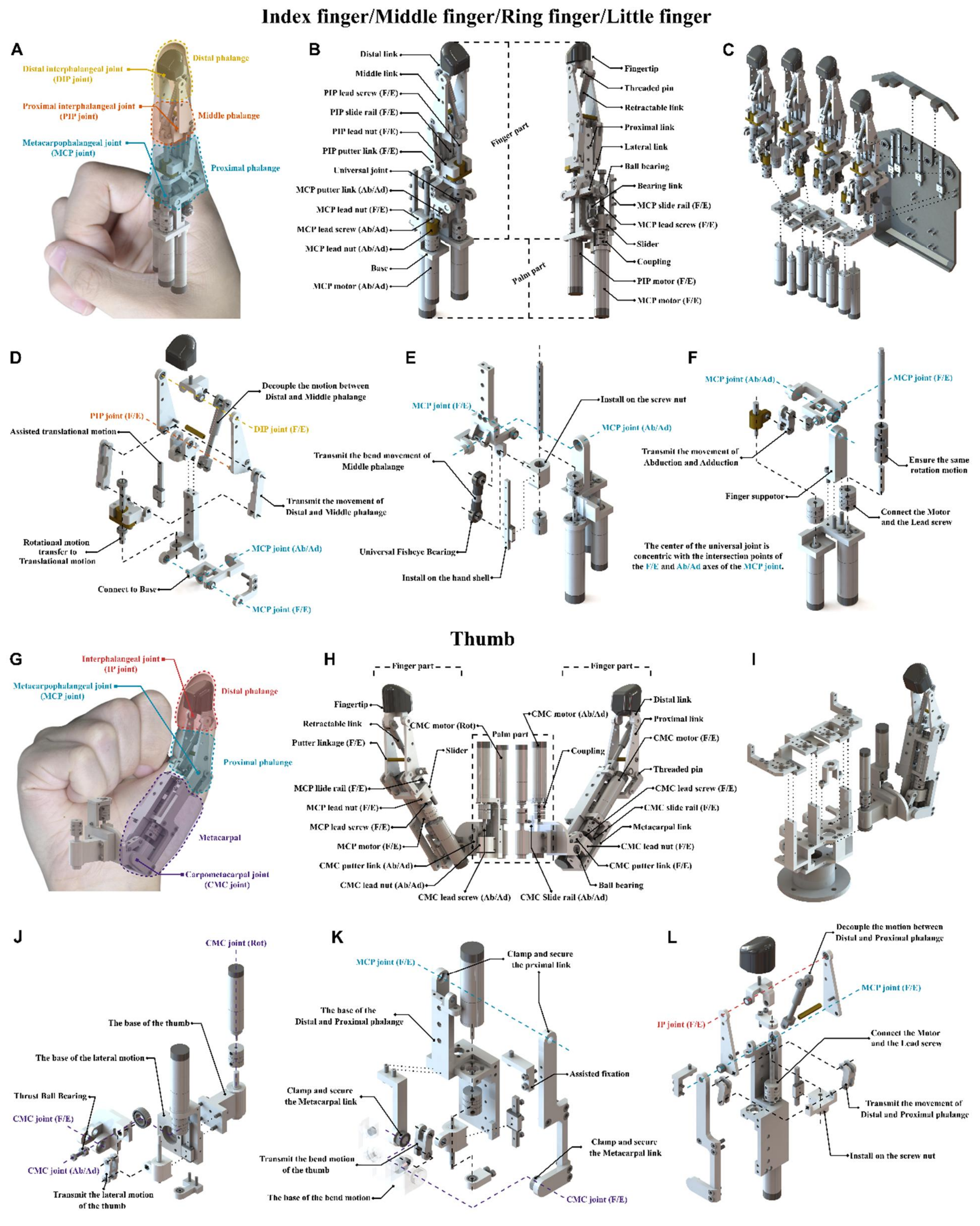

**Fig. 4. Mechanical implementation of Torch-Hand.** (A) Anthropomorphic correspondence between the four fingers of Torch-Hand and the human hand. (B) Mechanical structure of the four fingers module. (C) Assembly layout of the four-finger module within Torch-Hand. (D to F) Exploded views of the four fingers module. (G) Anthropomorphic correspondence between the thumb of Torch-Hand and the human hand. (H) Mechanical structure of the thumb module. (I) Assembly layout of the thumb module within Torch-Hand. (J to L) Exploded views of the thumb module.

The linkage transmission is further adapted to accommodate the spatial motion required by the finger joints. The MCP (F/E) motion is realized through a spatial four-bar mechanism, in which the driving link is equipped with spherical rod-end bearings at both ends to accommodate changes in joint orientation during motion. For PIP (F/E), power is transmitted through a universal joint whose motion center is aligned with the intersection of the MCP (F/E) and MCP (Ab/Ad) axes (Fig. 4F). This geometric alignment prevents the PIP transmission from interfering with MCP motions, allowing all DOF of the finger to operate either independently or in coordinated combination. As a result, the mechanical implementation preserves decoupled actuation while enabling coordinated multi-joint finger motion.

Beyond the four-finger transmission layout, the thumb requires a distinct mechanical architecture because opposition cannot be achieved by directly extending the same finger mechanism. In Torch-Hand, the thumb is therefore implemented as a mechanically differentiated module (Fig. 4H and 4I). At the carpometacarpal joint, the three rotational axes form an orthogonal local rotation basis, allowing thumb orientation to be decomposed into independent rotation, abduction/adduction and flexion/extension components (Fig. 4J and 4K). This arrangement minimizes parasitic coupling among CMC motions and enables compact, independent control of thumb approach direction, opposition posture and contact orientation. The MCP and IP joints are driven by local linkage transmissions that follow the same input-output principle used in the PIP and DIP joints of four fingers (Fig. 4L). This configuration allows the thumb to oppose the four fingers across an expanded workspace while preserving independently addressable joint motion.

Together, these implementation principles allow Torch-Hand to combine compact anthropomorphic morphology with independently actuated degrees of freedom, precise motion transmission, dexterous whole-hand manipulation and high load-bearing capability.

## KINEMATIC AND MECHANICAL PERFORMANCE OF TORCH-HAND

Torch-Hand is evaluated across complementary kinematic and mechanical performance dimensions required for dexterous humanoid manipulation. Kinematic performance is assessed through reachable workspace characterization, thumb opposition using the Kapandji test (73), and grasp diversity based on the 33-grasp taxonomy (74, 75). Mechanical performance is quantified through repeat positioning accuracy, grasping force measurement and payload capacity tests. These evaluations link the decoupled serial architecture and fully linkage-driven implementation to measurable dexterity, precision and load-bearing capability.

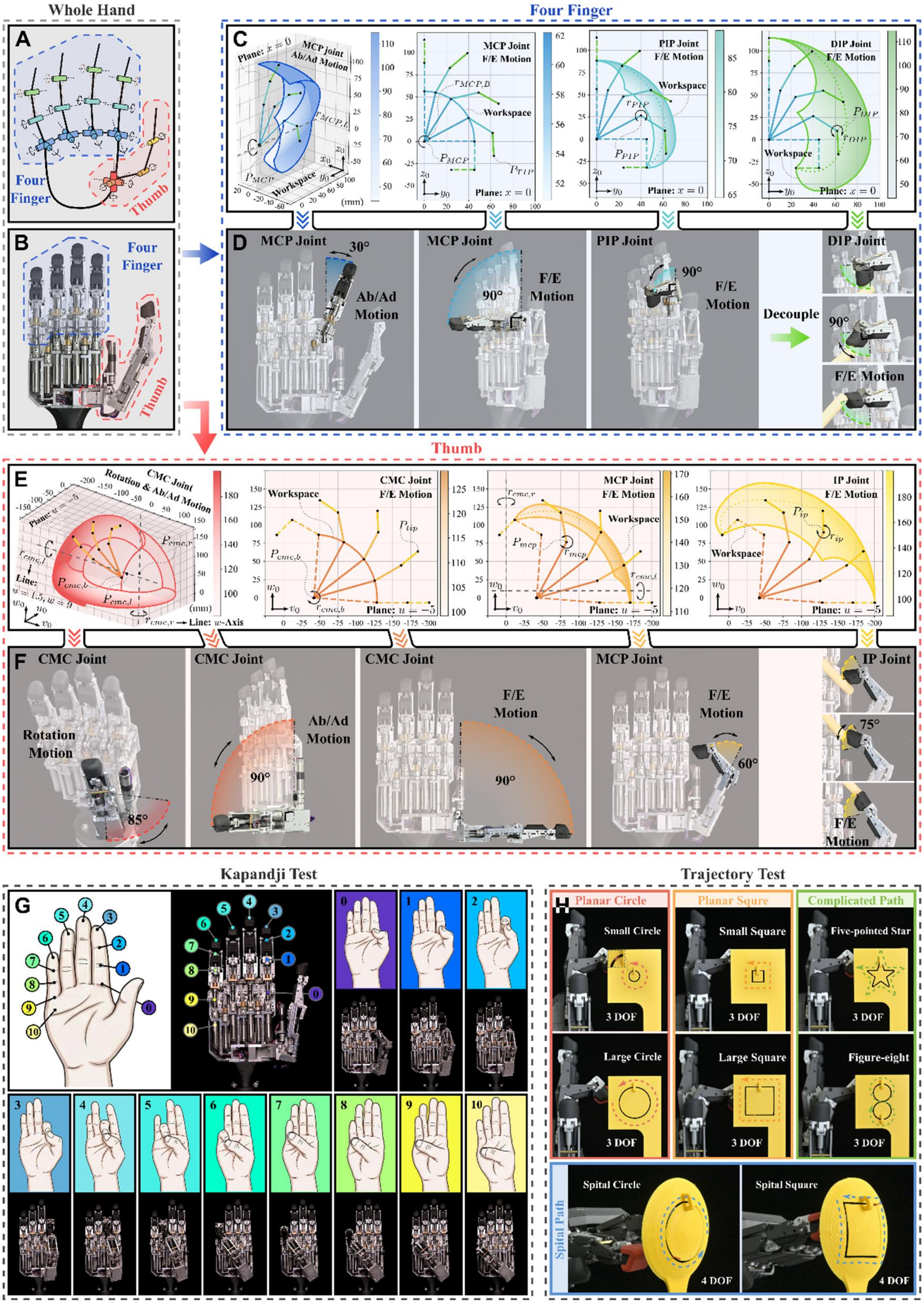

**Fig. 5: Kinematic range and fingertip workspace motion of Torch-Hand of Torch-Hand.** (A) Functional partitioning of the four-finger and thumb modules. (B) Kinematic assessment of the 21-DOF Torch-Hand. (C) Joint range of motion of the four fingers. (D) Fingertip workspace generated by four fingers motion. (E) Joint range of motion of the Thumb. (F) Fingertip workspace generated by thumb motion. (G) Planar and spatial trajectory tracking enabled by coordinated finger DOF. (H) Thumb opposition assessed by the Kapandji test.

To further examine coordinated fingertip dexterity, a slender stylus is attached to the fingertip and commanded to trace predefined trajectories through multi-joint coordination

The kinematic capability of Torch-Hand is first quantified by measuring joint range of motion and visualizing the corresponding fingertip workspaces. As shown in Fig. 5A and 5B, the 21 degrees of freedom are assessed across two functional modules: the four fingers and the thumb. For each of the four fingers, the MCP joint reached 90° in Ab/Ad and 30° in F/E, whereas the PIP and DIP joints each reached 90° in F/E (Fig. 5C). The F/E motions of the MCP, PIP and DIP joints define a planar fingertip workspace, which is laterally swept by MCP Ab/Ad into a three-dimensional reachable region with a smooth curved surface (Fig. 5D). This workspace supports both flexion-dominated grasp formation and lateral finger repositioning. For the thumb, the CMC joint achieved 85° in Rot, 90° in Ab/Ad and 90° in F/E, followed by 60° and 75° in F/E at the MCP and IP joints, respectively (Fig. 5E). The thumb workspace is generated by reorienting the F/E plane through CMC Rot and sweeping it through CMC Ab/Ad, producing an approximately quarter-spherical reachable space (Fig. 5F). This geometry expands thumb-finger opposition and improves coordination across diverse grasp postures.

(Fig. 5G). Six planar trajectories— small and large circles, small and large squares, a pentagram and a figure-eight path—are selected to span different motion demands, including continuous curvature control, scale variation, abrupt directional transitions, sharp vertices, curvature reversal and self-crossing motion. These planar tasks validate coordinated motion among the three flexion/extension joints of the four fingers. Two additional spatial trajectories assessed full-DOF coordination, requiring MCP Ab/Ad to operate with the three flexion/extension joints to generate three-dimensional fingertip motion. Successful execution across these trajectories shows that independently actuated DOF can be coordinated into smooth, continuous and task-specific fingertip motion, demonstrating the advantage of decoupled actuation for flexible multi-joint control.

Thumb opposition is evaluated using the Kapandji test, which assesses progressively more demanding thumb contacts across the fingers and palm (Fig. 5H). Torch-Hand successfully reaches all tested opposition positions, indicating effective thumb-finger and thumb-palm opposition enabled by the three-axis CMC mechanism and independently actuated thumb joints.

Grasp diversity is assessed using the 33-grasp taxonomy, a structured benchmark of human grasp postures spanning power, precision and intermediate grasp families (Fig. 6). Each grasp type is evaluated on both 3D-printed standard objects and real objects with corresponding shapes, allowing standardized posture reproduction and practical object adaptability to be assessed in parallel. Torch-Hand reproduces all 33 grasp types across both object sets. This complete repertoire shows that independently actuated DOF can be coordinated into diverse, stable hand postures and can support both force-oriented grasping and fine manipulation across object variations.

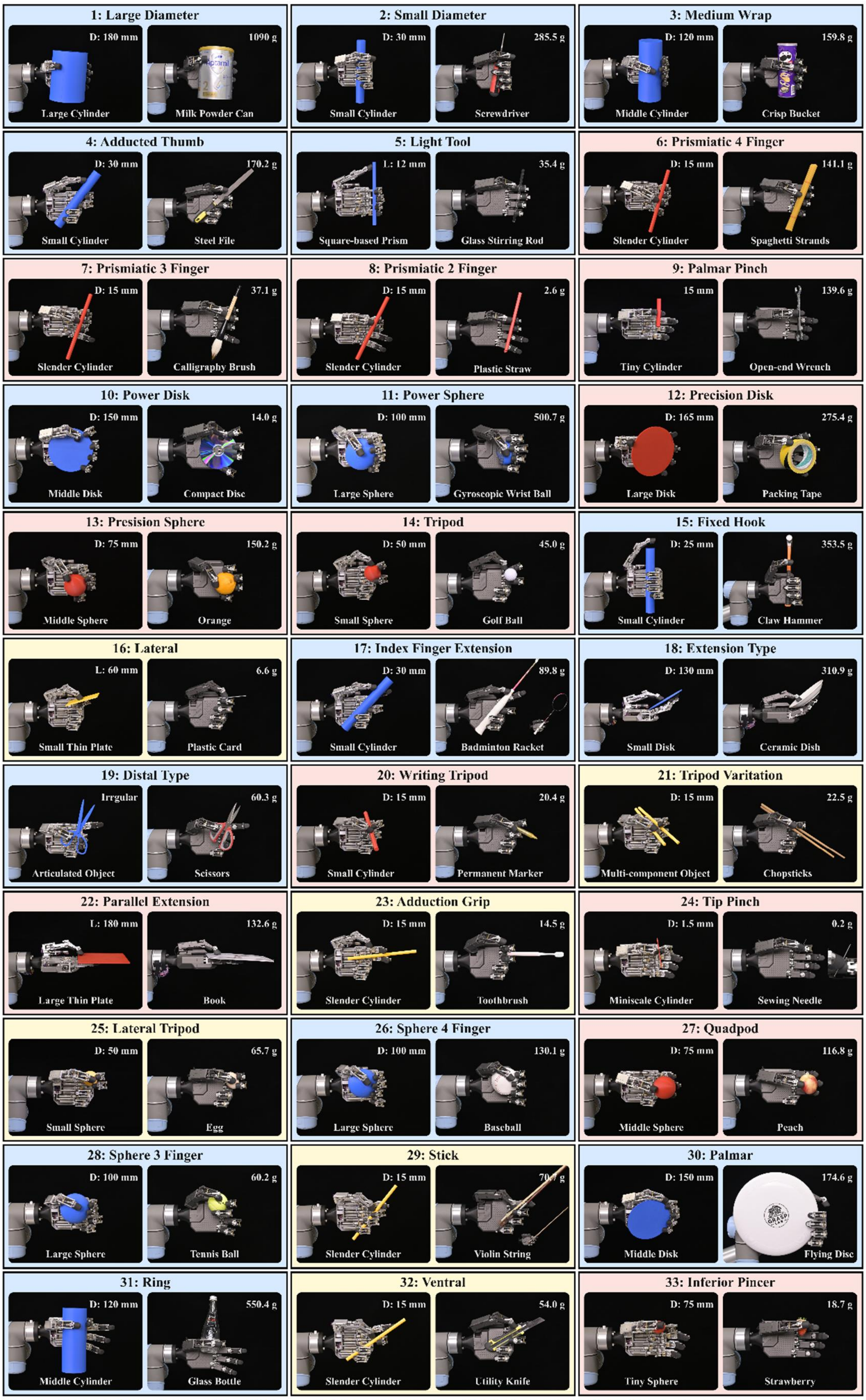

Blue Background: Power Grasp Yellow Background: Intermediate Grasp Red Background: Precision Grasp D: Diameter L: Edge Length

**Fig. 6. Grasp diversity across standardized and real-object 33-grasp taxonomy.**

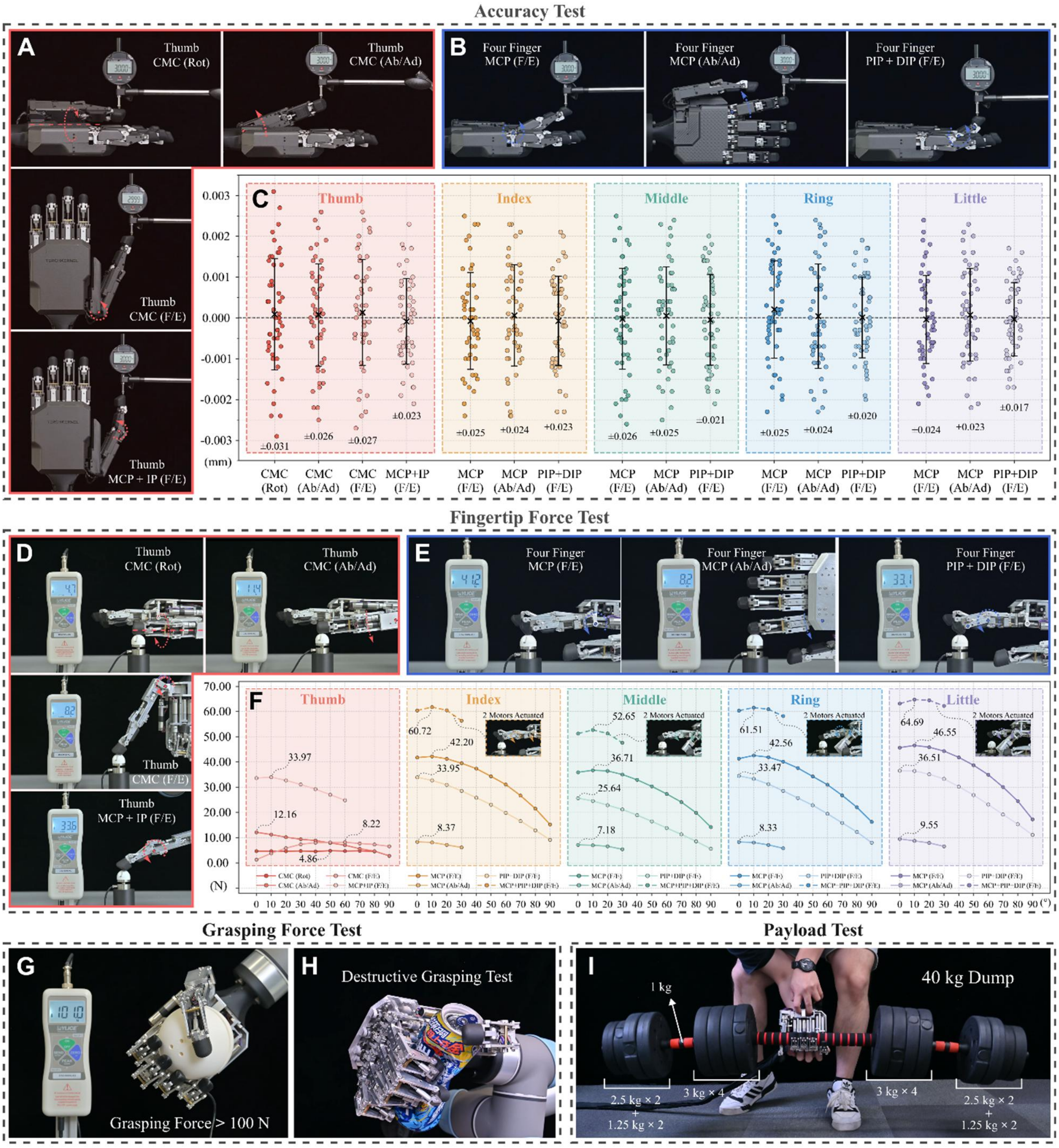

**Fig. 7: Mechanical precision and high-load grasping performance of Torch-Hand.** (A and B) Repeat positioning accuracy measurement of the four fingers and thumb. (C) Distribution of endpoint positioning errors across tested joints. (D and E), Fingertip force measurement of the four fingers and thumb. (F) Joint-angle-dependent maximum average force output and average fingertip force under coordinated MCP-PIP actuation. (G and H) Whole-hand grasping force test. (I) Load-bearing demonstration with a 40 kg dumbbell.

Following the kinematic evaluation, Torch-Hand is assessed for mechanical precision and load-bearing performance, including repeat positioning accuracy, grasping force and payload capacity. Repeat positioning accuracy is quantified separately for the four fingers and the thumb (Fig. 7A and 7B). The four fingers achieved a mean positioning error of 0.0111 mm, with errors bounded within ±0.026 mm, whereas the thumb achieved a mean positioning error of 0.0115 mm, with errors bounded within ±0.031 mm (Fig. 7C). These results show that the screw-driven slider and rigid linkage transmissions provide highly repeatable endpoint positioning across both functional modules.

Grasping force measurements quantified the force transmission capability of the fully linkage-driven hand (Fig. 7D and 7E). Single-joint actuation is first tested to establish the local force output of individual transmission pathways. In the four fingers, coordinated MCP-PIP actuation further increased fingertip force by allowing multiple flexion joints to contribute simultaneously to contact generation, producing average fingertip forces of 60.72 N, 52.65 N, 61.51 N and 64.69 N for the index, middle, ring and little fingers, respectively (Fig. 7F). At the whole-hand level, Torch-Hand generated a grasping force exceeding 100 N on a grip-force ball and crushed an aluminum can, demonstrating that local force outputs can be integrated into strong grasping actions (Fig. 7G and 7H). The hand also lifted a 40 kg dumbbell, confirming that the screw-linkage transmission and self-locking structure support high load-bearing manipulation after grasp formation (Fig. 7I). These results show that Torch-Hand combines repeatable force transmission, strong whole-hand grasping and substantial payload capacity.

These evaluations show that Torch-Hand combines kinematic versatility with mechanical robustness. Although its joints are mechanically decoupled, they can be coordinated into coupled task-level motions, enabling diverse grasp formation, repeatable transmission, strong force output and high load-bearing capability within a compact anthropomorphic architecture.

## TASK-LEVEL MANIPULATION DEMONSTRATIONS OF TORCH-HAND

To assess functional manipulation beyond isolated performance metrics, Torch-Hand is tested using a 3×3 task matrix. Figure 8 summarizes this task-level evaluation across two axes: manipulation dexterity and interaction load. The horizontal axis represents increasing manipulation demand, from dexterous fingertip operation to coordinated object handling and forceful tool-oriented manipulation, whereas the vertical axis represents increasing interaction load. The resulting nine tasks cover representative manipulation scenarios ranging from fine fingertip operation to high-load tool use.

Representative tasks further illustrate how the mechanical and kinematic advantages of Torch-Hand translate into scenario-level functionality. In a household context, the hand opens a lipstick cap and twists the lipstick body, requiring thumb-finger opposition, fingertip repositioning and coordinated multi-joint rotation. This demonstration shows that mechanically decoupled finger DOF can be actively coordinated into coupled multi-joint motion patterns. For service-oriented precision operation, Torch-Hand threads a 0.35 mm nylon line through a 0.5 mm needle eye, demonstrating fine fingertip control and multi-finger coordination under tight geometric constraints. In a laboratory workflow, pipette-based dilution required precise fingertip positioning and controllable force application to regulate droplet release. Industrial capability is demonstrated by extracting a nail from a wooden board using a claw hammer, showing that the same anthropomorphic platform can sustain high-force tool interaction with stable whole-hand grasping.

Successful execution across this task matrix shows that Torch-Hand integrates mechanically decoupled joint control with coordinated multi-joint action across diverse manipulation demands. Independently addressable DOF provide precise endpoint regulation, while coordinated activation enables coupled motion patterns for complex manipulation. This combination allows the hand to transition from delicate fingertip tasks to high-load tool use while maintaining thumb-finger opposition, motion coordination and stable forceful grasping.

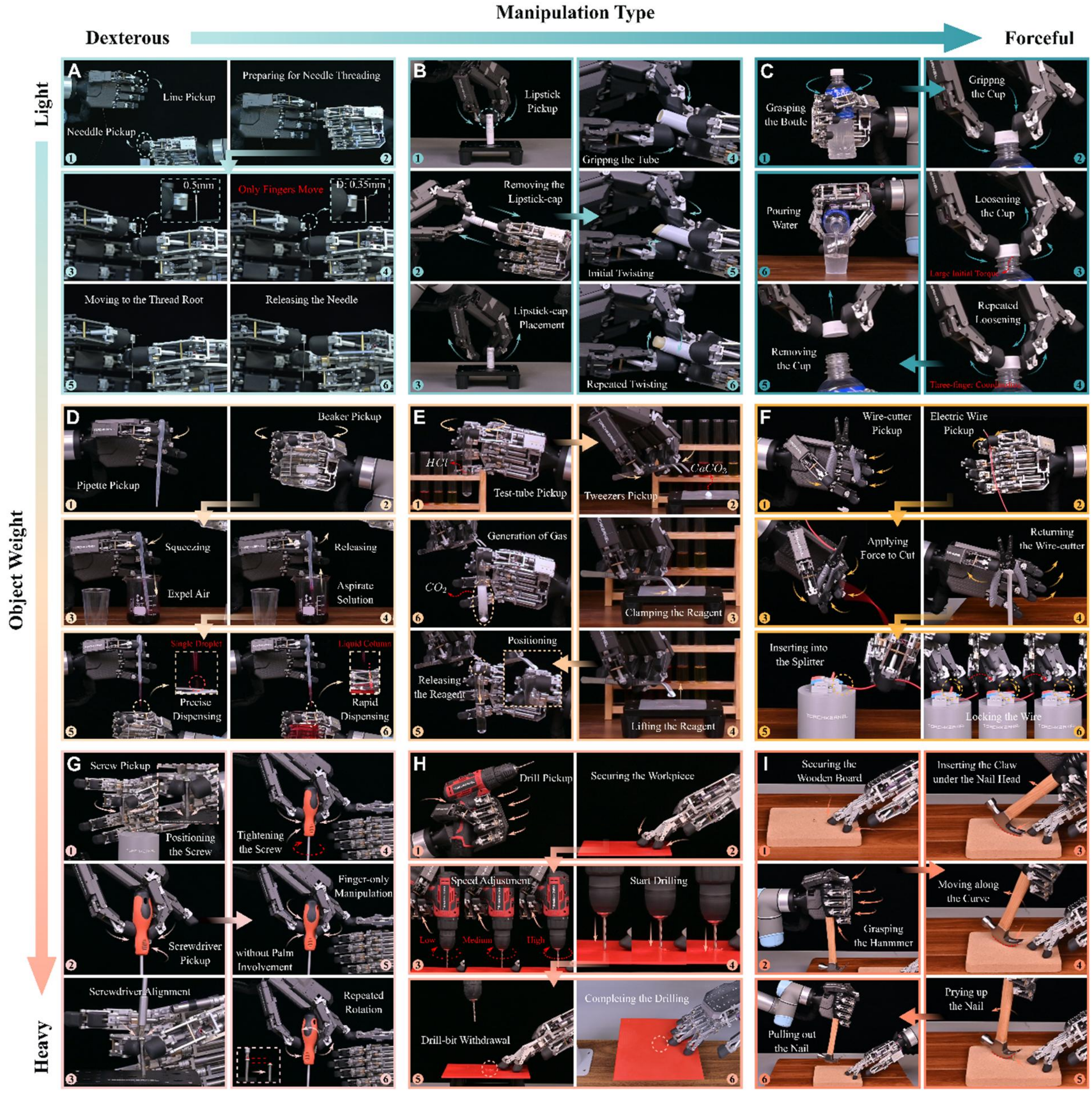

**Fig. 8: Task-level dexterous manipulation across dexterity and load demands.** (A) Needle threading with a nylon line. (B) Lipstick-cap opening and lipstick twisting. (C) Unscrewing of a tightly sealed bottle cap. (D) Pipette-based dilution and droplet regulation. (E) Calcium carbonate handling for carbon dioxide generation. (F) Wire cutting and insertion into a wire splitter. (G) Controlled screw tightening with stable rotational motion. (H) Electric drilling with stable tool grasp. (I) Nail extraction from a wooden board using a claw hammer.

## DISCUSSION

This study presents Torch-Hand, a fully linkage-driven humanoid dexterous hand that reconciles high-DOF dexterity, motion transparency and high-load manipulation within a compact anthropomorphic architecture. Torch-Hand preserves the kinematic richness of the human hand through a 21-DOF structure, while concentrating all actuators within the palm and transmitting motion to lightweight fingers through compact screw-linkage mechanisms. This palm-integrated actuation layout reduces distal inertia, preserves anthropomorphic finger proportions and leaves space within the digits for linkage transmission and potential sensory integration. The use of elementary yet mechanically stable linkage and screw-driven units provides a robust and analytically traceable transmission basis for precise motion and forceful interaction. Organizing each finger as a serial robotic chain further simplifies the kinematic relationship from actuator input to fingertip motion, improving motion transparency and control tractability. More importantly, the results show that mechanical decoupling and coordinated dexterity are not contradictory. Mechanically decoupled finger DOF can be actively coordinated into coupled multi-joint motion patterns, enabling precise endpoint regulation, complex fingertip trajectories, diverse grasp configurations and force-demanding tool manipulation within a compact hand-scale form factor.

A central implication of Torch-Hand is that mechanical decoupling can improve controllability without limiting dexterous behavior. The serial finger configuration assigns each functional DOF to an independently addressable transmission pathway, yielding a transparent actuator-joint-fingertip mapping and reducing unintended motion propagation between joints. This architecture provides local control authority for precise endpoint regulation, as reflected by the repeatability and trajectory-tracking results, while the complete 33-grasp repertoire and task matrix show that the same DOF can be combined into complex fingertip paths, diverse hand postures and functional manipulation across different interaction demands. This distinction between mechanical decoupling and behavioral coordination is important for high-DOF robotic hands: dexterity does not require permanently coupled joints, but rather independently controllable joints that can be coordinated when needed.

The fully linkage-driven implementation addresses a central challenge in humanoid hand design: achieving dexterous motion without sacrificing mechanical output. By relocating actuators to the palm and transmitting power through rigid screw-linkage pathways, Torch-Hand keeps the fingers lightweight while preserving deterministic force and motion transmission. Rigid linkages reduce uncertainty associated with compliant or cable-based power transfer, whereas screw-driven sliders provide mechanical gain, positional stability and self-locking. This combination enables fine motion regulation and high-force interaction to coexist within the same compact hand architecture. This balance is reflected in the mechanical performance results: endpoint positioning errors are bounded within ±0.026 mm for the four fingers and ±0.031 mm for the thumb, while the maximum measured single-finger output force reached 64.69 N. Crushing an aluminum can, generating more than 100 N of whole-hand grasping force and lifting a 40 kg dumbbell further demonstrate that high-load manipulation can be achieved without bulky distal actuators or purely underactuated grasping strategies.

The task-level demonstrations further show that the proposed architecture provides a practical hardware basis for embodied manipulation rather than only isolated performance gains. The 3×3 task matrix spans dexterous fingertip operation, object handling and forceful tool use under increasing interaction loads, with representative tasks such as needle threading, pipette-based droplet regulation, lipstick twisting and claw-hammer nail extraction requiring distinct combinations of precision, opposition, coordination, grasp stability and force output. The ability to operate across household, service, laboratory and industrial contexts without changing the mechanical configuration highlights the generality of the hand platform. For embodied intelligence, this is a critical advantage: general-purpose manipulation requires not only perception and control policies, but also a body that can deliver precise, strong and coordinatable interaction with diverse objects and tools. Torch-Hand therefore offers a mechanically capable substrate for future integration with visual-tactile sensing, closed-loop control and learning-based manipulation strategies.

Torch-Hand offers a mechanically capable and extensible hardware substrate for future embodied manipulation systems, on which perception, feedback and learning can be further integrated. Its mechanically transparent and decoupled architecture provides a structured control space: actuator inputs, joint motions and fingertip behaviors remain analytically traceable, while human hand joint motions can be mapped more directly to robotic hand postures for teleoperation. The modular palm-finger construction further improves manufacturability, maintenance and extensibility, supporting lower-cost production, component-level repair and the integration of additional sensing or actuation modules. Building on these hardware advantages, visual perception, distributed tactile sensing and force feedback could substantially enhance Torch-Hand's sensorimotor capability. The independently addressable DOF and clear actuator-joint-fingertip mapping provide a favorable basis for closed-loop regulation, while added sensory feedback would enable adaptive control of contact force, slip response, grasp posture and fragile-object interaction. Such a sensor-enhanced platform could extend Torch-Hand's current strengths—high-DOF dexterity, precise transmission and high-load manipulation—toward more autonomous and adaptive operation in unstructured environments. These capabilities are particularly relevant for medical assistance, rehabilitation training, teleoperation, human augmentation and hazardous-environment manipulation, where human-like dexterity must be combined with reliable physical interaction. Future deployment should also include systematic evaluation of long-term durability, fatigue resistance and reliability under dynamic tool use and sustained high-load interaction.

Torch-Hand shows that humanoid hand design need not trade dexterity for mechanical strength or control tractability. By combining independently addressable DOF, serial motion transparency and screw-linkage force transmission, the hand provides a compact hardware route for precise, coordinatable and high-load manipulation. This architecture suggests a scalable foundation for embodied robotic systems in which mechanical design actively supports, rather than merely executes, intelligent physical interaction.

## MATERIALS AND METHODS

### MECHANICAL DESIGN AND FABRICATION

Torch-Hand is implemented as a modular, palm-integrated humanoid hand platform. The four finger modules and the thumb module are fabricated as replaceable linkage assemblies and connected to the palm through modular mechanical interfaces. The main palm frame and dorsal cover are fabricated from titanium alloy (TC4), whereas the phalanges and linkage components are fabricated from aluminum alloy (6061). Flexible fingertip pads are made from soft rubber material (WeNext Ltd.) to provide sufficient contact friction during grasping and manipulation. The internal mechanisms are protected by three-dimensional printed outer shells (ABS, Bambu Lab Ltd.).

The actuation system is integrated within the palm. MCP abduction/adduction of the four fingers is driven by 10-mm-diameter low-power motors (JCCM1039, JIECANG Ltd.), whereas thumb CMC rotation is driven by a 10-mm-diameter high-power motor (JCCM1045, JIECANG Ltd.). MCP flexion/extension of the four fingers is driven by 12-mm-diameter high-power motors (JCCM1254, JIECANG Ltd.), and the remaining degrees of freedom are driven by 12-mm-diameter low-power motors (JCCM1245, JIECANG Ltd.). Each motor is connected to the corresponding screw through a coupling (CI-10×12, Xingda Inc.). Ball screws (NSG0401, KSS Ltd.) are used for the MCP flexion/extension joints of the four fingers, whereas trapezoidal screws (MG4, KSS Ltd.) are used for the remaining joints. Each screw-driven transmission is assembled with a miniature guide rail (MGNR3, HIWIN Ltd.) and slider block (MGN3C, HIWIN Ltd.). The slider output is connected to the corresponding linkage unit to convert linear motion into joint rotation. Revolute joints in the linkage mechanisms are supported by deep-groove ball bearings from the MR series (NSK Ltd.) or sleeve bearings from the PTFE series (MFG Ltd.) to reduce friction and improve motion precision. After fabrication, each finger and thumb module is inspected independently before being mounted to the palm module.

### MODULAR ASSEMBY PROCEDURE

Torch-Hand is assembled using a modular palm-finger workflow, in which the four finger modules, thumb module and palm module are prepared separately before final integration. This procedure allowed each transmission chain to be inspected independently before whole-hand assembly.

For each of the four finger modules, the proximal, middle and distal phalanges are installed sequentially, followed by the fingertip pads. The push rods and linkage components on both sides of the PIP joint are inspected for symmetry, and the DIP compressible-link mechanism is manually checked to verify PIP-DIP coordination during free motion and decoupling under compression. The thumb module is assembled by first installing the CMC Ab/Ad and CMC Rot mechanisms and fixing the CMC Rot mechanism to the thumb base. The CMC F/E, MCP F/E and IP F/E linkage mechanisms are then installed sequentially. After assembly, the thumb module is manually actuated through its full range of motion to confirm smooth transmission and absence of mechanical interference.

The palm module is prepared by fixing the four-finger actuators to the palm base and attaching the dorsal cover. The four finger modules and thumb module are then mounted to the palm through modular mechanical interfaces and couplings. During palm-finger integration, the four-finger modules are positioned such that the intersection of the MCP F/E and MCP Ab/Ad axes coincided with the center of the universal joint, minimizing transmission interference among the finger DOF. Electrical wires are routed through predefined palm channels to avoid interference with moving transmission components. Before experimental testing, each DOF is actuated independently and in selected multi-DOF combinations to verify transmission direction, joint range and smooth motion. The detailed assembly workflow is shown in Fig. S2.

## ACTUATION, LINKAGE TRASMISSION AND SERIAL KINEMATIC MODELING

For each actuated degree of freedom, the motor output is modeled as a local actuator-slider-linkage transmission pathway. The angular displacement of the iii-th motor, denoted by $\psi_i$, is first converted into the linear displacement $s_i$ of the corresponding active slider through the screw transmission. For a screw with lead $p_i$, the slider displacement is calculated as

$$s_i = \frac{p_i}{2\pi}\psi_i \quad (3)$$

The slider displacement is then used as the input to the corresponding linkage mechanism. The joint angle $\theta_i$ is obtained from the geometric closure relationship of each slider-linkage unit,

$$\theta_i = f_i(s_i) \quad (4)$$

where $f_i(\cdot)$ denotes the mechanism-specific input-output function determined by the linkage geometry. The detailed derivation of the actuator-to-slider and slider-to-joint mappings is provided in Supplementary Eqs. S1-S27 and Fig. S1. The joint variables obtained from the local transmission models are then introduced into the serial kinematic formulation defined in the Results. Each finger is modeled as an open serial chain rooted at the palm-fixed frame and ending at the fingertip frame. Fingertip pose is calculated using the DH convention, in which the homogeneous transformations between adjacent joint frames were multiplied sequentially to obtain the fingertip frame relative to the palm frame. The fingertip position is extracted from the translational component of the final homogeneous transformation matrix. The complete transformation matrices, DH parameter definitions and fingertip pose extraction procedure are provided in Supplementary Eqs. S28-S32 and Supplementary Tables S2 and S3.

## EXPERIMENT SETUP FOR KINEMATIC PERFORMANCE EVALUATION

• Workspace. Fingertip workspaces are generated using the serial manipulator model combined with Monte Carlo sampling. Joint variables are randomly sampled within their measured motion ranges, with 50,000 sampled configurations for each of four fingers model and 100,000 sampled configurations for the thumb model. The corresponding fingertip positions are computed from the Denavit-Hartenberg model in the palm-fixed coordinate frame. The sampled fingertip positions form a reachable point cloud. Boundary points of the point cloud are then extracted to define the spatial envelope of the workspace, and a smooth surface is reconstructed from these boundary points for visualization. This procedure is applied consistently to the four fingers and the thumb.

• Trajectory-tracking. Experiments. A fingertip sleeve carrying a 1.5-mm-diameter rod is mounted on the fingertip, and the rod is guided along predefined planar and spatial trajectories with a path width of 2 mm. For each trajectory, the joint-angle sequence is generated in advance and sent to the hand controller during execution. Six planar trajectories, including large and small circles, large and small squares, a pentagram and a figure-eight path, are commanded at 20 Hz to assess coordination among the MCP F/E, PIP F/E and DIP F/E joints. Two spatial trajectories are commanded at 30 Hz to evaluate full-DOF coordination, with MCP Ab/Ad actuated together with the three flexion/extension joints. Successful tracking is defined as continuous motion of the fingertip-mounted rod along the prescribed path without leaving the 2-mm trajectory width or causing visible motion interruption.

• Kapandji test. For the Kapandji test, prescribed contact regions are defined on the four fingers and palm in the palm-fixed coordinate frame. For each opposition target, the posture command includes the thumb joints and the corresponding interacting finger joints. Joint angles are obtained from the kinematic relationship between the thumb and target finger chains within their measured joint limits, and are then executed on the physical hand. Small joint-angle offsets are applied when necessary to compensate for assembly tolerance and contact compliance. The calibrated thumb-finger joint angles are saved as the posture command for each target. During testing, these commands are executed sequentially, and a target is considered successful when the thumb tip contacts the prescribed region without external assistance or hand repositioning.

• 33-grasp taxonomy. For each grasp type, a human subject performs the corresponding grasp on a representative object while wearing a calibrated data glove. The glove is calibrated using reference hand postures before data collection, and the joint angles of the human hand are recorded during stable grasp formation. The recorded human joint angles are mapped to the corresponding Torch-Hand joints according to anatomical joint correspondence. Because the human hand and Torch-Hand differ in size, the objects used for robotic grasping are scaled according to their geometric ratio, preserving the relative object-hand relationship. The mapped joint-angle command is then executed on Torch-Hand, and small joint-angle offsets are applied only when necessary to compensate for object contact, assembly tolerance and compliance. A grasp is considered successful when the object is held stably for 10 s without external support, visible slip or loss of the intended grasp configuration. The demonstrations are provided in Supplementary Videos S1-S5.

## EXPERIMENT SETUP FOR MECHANICAL PERFORMANCE EVALUATION

• Repeat positioning accuracy is measured separately for all fingers. During each trial, the tip of a high-resolution dial indicator probe is positioned 3.000 mm below the theoretical terminal position of the fingertip. The tested finger is driven from its neutral posture to the terminal posture, and the indicator reading is recorded after the measurement stabilizes. Each joint is tested for 50 repeated trials. The endpoint positioning error is calculated as the difference between the stabilized indicator reading and the reference value of 3.000 mm. The mean error, standard deviation and maximum bounded error are calculated for each tested finger. Detailed repeatability data and the experimental procedure are provided in Supplementary Table S4 and Supplementary Video S5.

• Fingertip force. Fingertip force is measured using a horizontally placed force sensor (500 N, Yilice Inc.). The contact surface between the sensor and the fingertip is designed as a spherical interface to provide consistent contact during loading. For each tested joint, the joint angle is increased from 0° to its measured upper limit in 10° increments. At each angle, the tested finger presses the sensor vertically from above. During the pressing phase, the corresponding actuator is operated in current-loop mode, and the motor current is increased to the preset upper limit for that actuator. After the fingertip force reading stabilizes, the stabilized value is recorded as the fingertip force. The actuator is then switched to position-loop control to return the finger to its initial position, completing one measurement cycle. Each joint-angle condition is tested 10 cycles. Single-joint actuation is first tested to quantify the local force output of individual transmission pathways. For the four fingers, coordinated MCP-PIP actuation is also tested to evaluate the combined force output generated by simultaneous flexion of the two joints. Detailed force measurements and the experimental procedure are provided in Supplementary Table S5 and Supplementary Video S6.

• Force & Load. Whole-hand force output is evaluated using a grip-force ball with an embedded force sensor. During testing, Torch-Hand compresses the ball with all five digits, and the stabilized peak force is recorded from the sensor output. A wedge-shaped stopper is fixed to the palm to prevent ball slippage during compression (Fig. S3). Destructive grasping is evaluated qualitatively by crushing an aluminum can, and payload capacity is tested by lifting a 40 kg dumbbell after forming a stable grasp around the handle. A payload trial is considered successful when the grasp is maintained without visible slip, structural failure or loss of grasp posture during lifting. All demonstrations are recorded in video S7.

### TASK-LEVEL MANIPULATION DEMONSTRATIONS

Left- and right-hand versions of Torch-Hand are mounted on robotic arms (UR-5, Universal Robots Ltd.) according to the requirements of each task. For arm motion, the required end-effector poses are obtained in advance through teleoperation and replayed during the task demonstration. For hand motion, a human demonstrator wears a data glove and performs the corresponding manipulation, while the human hand joint-angle sequence is recorded at 20 Hz. The recorded joint angles are used to reconstruct the fingertip trajectory of each digit using a human-hand kinematic model, and the trajectories are then expressed in the Torch-Hand palm-fixed coordinate frame. The Torch-Hand joint-angle sequence is obtained by solving inverse kinematics for each digit from the retargeted fingertip trajectories.

During each demonstration, the UR-5 arm first moves Torch-Hand to the recorded task pose, and the hand joint-angle sequence is then sent to the controller at 20 Hz to reproduce the required grasping or manipulation motion. A task is considered successful when Torch-Hand completes the intended functional outcome without external assistance after initial object or tool placement, while maintaining the required grasp or contact state without visible slip, loss of posture or task interruption. All task demonstrations are recorded by video, and representative demonstrations are provided in Supplementary Video S8.

## Supplementary Materials

Supplementary Methods
Figs. S1 to S2
Tables S1 to S6
Movies S1 to S8

**Acknowledgments:**

**Author contributions:**

Conceptualization: J.L., H.D.
Methodology: J.L.
Investigation: J.L., J.W.
Validation: J.L., J.W.
Visualization: J.L., H.G.
Funding acquisition: G.L., H.D.
Supervision: G.L.
Writing - original draft: J.L., Y.W.
Writing - review & editing: J.L., H.G., Y.W., H.D.

**Competing interests:** Authors declare that they have no competing interests.

**Data and materials availability:** All data needed to evaluate the conclusions in the paper are present in the paper and/or the Supplementary Materials.

**Title:** A fully articulated humanoid dexterous hand with human-level joint structure enabled by linkage-based actuation

**Authors:**
Jihao Li[1,2], Jianguo Wang[1,2], Haotian Guo[1], Yanzhe Wang[1], Guodong Lu[2], Huixu Dong[1,2]*

**Supplementary Materials**

Supplementary Methods
Figs. S1 to S2
Tables S1 to S6
Movies S1 to S8

**Supplementary Methods**

We can use the position of the slider in a crank-slider mechanism to calculate the angles between the links, and then calculate the bending angles of each finger joint, which can then be used for calculations in DH coordinates.

Use $\boldsymbol{s_{MCP,L}}$ to calculate $\theta_{MCP,L}$, and when the angle $\theta_{MCP,L}$ is 0, the slider position is located at $\boldsymbol{s_{MCP,L}}_0$ and $\theta'_{MCP,L}$ is $\theta_{MCP,L0}$.

$$|\boldsymbol{l_G}| = \sqrt{d_{MCP,L}{}^2 + |\boldsymbol{s_{MCP,L}}{}^2|} \tag{S1}$$

$$\theta'_{MCP,L} = \arcsin\frac{d_{MCP,L}}{|\boldsymbol{l_G}|} + \arccos\frac{(|\boldsymbol{l_{A1}}|^2+|\boldsymbol{l_{A3}}|^2)+|\boldsymbol{l_G}|^2-|\boldsymbol{l_{T1}}|^2}{2|\boldsymbol{l_{T1}}||\boldsymbol{l_G}|},\ \theta'_{MCP,L} \in [0,\pi] \tag{S2}$$

$$\theta_{MCP,L} = \theta_{MCP,L} - \theta_{MCP,L0} \tag{S3}$$

Use $\boldsymbol{s_{MCP,B}}$ to calculate $\theta_{MCP,B}$, and when the angle $\theta_{MCP,B}$ is 0, the slider position is located at $\boldsymbol{s_{MCP,B0}}$ and $\theta'_{MCP,B}$ is $\theta_{MCP,B0}$.

$$O\ (0,0,0),\ O_x\ (|\boldsymbol{l_{B1}}|\sin\theta'_{MCP,B}\sin\theta_L,\ -|\boldsymbol{l_{B1}}|\cos\theta'_{MCP,B},\ |\boldsymbol{l_{B1}}|\sin\theta'_{MCP,B}\cos\theta_{MCP,L}),\ O_y\ (0, d_{MCP,B}, |\boldsymbol{s_{MCP,B}}|) \tag{S4}$$

$$\left|\overrightarrow{OO_y} - \overrightarrow{OO_x}\right| = |\boldsymbol{l_{T2}}| \tag{S5}$$

$$|\boldsymbol{l_{T2}}|^2 = (|\boldsymbol{l_{B1}}|\sin\theta'_{MCP,B}\sin\theta_{MCP,L} - 0)^2 + (-|\boldsymbol{l_{B1}}|\cos\theta'_{MCP,B} - d_{MCP,B})^2 + (|\boldsymbol{l_{B1}}|\sin\theta'_{MCP,B}\cos\theta_{MCP,L} - |\boldsymbol{s_{MCP,B}}|)^2 \tag{S6}$$

$$\theta'_{MCP,B} = \mathrm{atan2}(-|\boldsymbol{s_{MCP,B}}|, d_{MCP,B}) \pm \arccos\frac{|\boldsymbol{l_{T2}}|^2-|\boldsymbol{l_{B1}}|^2-d_{MCP,B}{}^2-|\boldsymbol{s_{MCP,B}}|^2}{2\sqrt{d_{MCP,B}{}^2+|\boldsymbol{s_{MCP,B}}|^2\cos^2\theta_{MCP,L}}},\ \theta'_{MCP,B} \in [0,\pi] \tag{S7}$$

The function $\mathrm{atan2}(x,y)$ denotes the two-argument inverse tangent, which returns the polar angle of the vector $(x,y)$ measured from the positive $x$-axis. Unlike the single-argument inverse tangent $\arctan(y/x)$, $\mathrm{atan2}(y/x)$ uniquely determines the angle over the full range $(-\pi,\pi]$ by explicitly accounting for the signs of both $x$ and $y$, thereby resolving the quadrant ambiguity and remaining well-defined when $x = 0$.

Use $\boldsymbol{s_{PIP}}$ to calculate $\theta_{PIP}$, and when the angle $\theta_{PIP}$ is 0, the slider position is located at $\boldsymbol{s_{PIP0}}$ and $\theta'_{PIP}$ is $\theta_{PIP}$.

$$|\boldsymbol{l_H}| = \sqrt{{d_{PIP}}^2 + |\boldsymbol{s_{PIP}}|^2} \tag{S8}$$

$$\theta'_{PIP} = \arcsin\frac{d_{PIP}}{|\boldsymbol{l_H}|} + \arccos\frac{|\boldsymbol{l_{B2}}|^2+|\boldsymbol{l_H}|^2-|\boldsymbol{l_{T3}}|^2}{2|\boldsymbol{l_{B2}}|l_H} \tag{S9}$$

$$\theta_{PIP} = \theta'_{PIP} - \theta_{PIP0} \tag{S10}$$

When the PIP joint is bent to its target position, the DIP joint can still be bent using the retractable link, $|\boldsymbol{s_{DIP0}}|$ is the original length of the retractable link, and $|\boldsymbol{s_{DIP}}|$is the actual length of it. Thus, we can use $|\boldsymbol{s_{DIP}}|$and $\theta_{PIP}$ to calculate $\theta_{DIP}$.

$$|\boldsymbol{l_K}| = \sqrt{|\boldsymbol{l_{PIP}}|^2 + |\boldsymbol{l_{B4}}|^2 - 2\cos(\Phi - \theta_{PIP})|\boldsymbol{l_{PIP}}|\,|\boldsymbol{l_{B4}}|} \tag{S11}$$

$$\theta'_{DIP} = \arccos\frac{|\boldsymbol{l_{B3}}|^2+|\boldsymbol{l_K}|^2-|\boldsymbol{s_{DIP}}|^2}{2|\boldsymbol{l_{B3}}||\boldsymbol{l_K}|}, \theta'_{DIP} \in [0,\pi] \tag{S12}$$

$$\Lambda = \arccos\frac{|\boldsymbol{l_{PIP}}|^2+|\boldsymbol{l_K}|^2-|\boldsymbol{l_{B4}}|^2}{2|\boldsymbol{l_{PIP}}||\boldsymbol{l_K}|}, \Lambda \in [0,\pi] \tag{S13}$$

$$\theta_{DIP} = \theta'_{DIP} + \Gamma - \Lambda - \pi \tag{S14}$$

Similarly, we can use a similar method to calculate the DH parameter of the thumb.

Use $\boldsymbol{s_{cmc,l}}$ to calculate $\theta_{cmc,l}$, and when the angle $\theta_{cmc,l}$ is 0, the slider position is located at $\boldsymbol{s_{cmc,l0}}$ and $\theta'_{cmc,l}$ is $\theta_{cmc,l0}$.

$$|\boldsymbol{l_g}| = \sqrt{{d_{cmc,l}}^2 + |\boldsymbol{s_{cmc,l}}|^2} \tag{S15}$$

$$\begin{cases} \theta'_{cmc,l} = \arccos\frac{|\boldsymbol{l_a}|^2+|\boldsymbol{l_g}|^2-|\boldsymbol{l_{t1}}|^2}{2|\boldsymbol{l_a}||\boldsymbol{l_g}|} - \arctan\frac{|\boldsymbol{s_{cmc,l}}|}{d_{cmc,l}}, O_a \text{ is below } line\ l_1 \\ \theta'_{cmc,l} = \arccos\frac{|\boldsymbol{l_a}|^2+|\boldsymbol{l_g}|^2-|\boldsymbol{l_{t1}}|^2}{2|\boldsymbol{l_a}||\boldsymbol{l_g}|} + \arctan\frac{|\boldsymbol{s_{cmc,l}}|}{d_{cmc,l}}, O_a \text{ is above } line\ l_1 \end{cases}, \theta'_{cmc,l} \in [0,\pi] \tag{S16}$$

$$\theta_{cmc,l} = \theta'_{cmc,l} - \theta_{cmc,l0} \tag{S17}$$

Use $\boldsymbol{s_{cmc,b}}$ to calculate $\theta_{cmc,b}$, and when the angle $\theta_{cmc,b}$ is 0, the slider position is located at $\boldsymbol{s_{cmc,b0}}$ and $\theta'_{cmc,b}$ is $\theta_{cmc,b0}$.

$$|\boldsymbol{l_j}| = \sqrt{{d_{cmc,b}}^2 + |\boldsymbol{s_{cmc,b}}|^2} \tag{S18}$$

$$\theta'_{cmc,b} = \arccos\frac{d_{cmc,b}}{|\boldsymbol{l_j}|} + \arccos\frac{|l_{t2}|^2+|l_j|^2-|l_{b2}|^2}{2|l_{t2}||l_j|}, \theta'_{cmc,b} \in [0,\pi] \tag{S19}$$

$$\theta_{cmc,b} = \theta'_{cmc,b} - \theta_{cmc,b0} \tag{S20}$$

Use $\boldsymbol{s_{mcp}}$ to calculate $\theta_{mcp}$, and when the angle $\theta_{mcp}$ is 0, the slider position is located at $\boldsymbol{s_{mcp0}}$ and $\theta'_{mcp}$ is $\theta_{mcp0}$.

$$|\boldsymbol{l_h}| = \sqrt{{d_{mcp}}^2 + {\boldsymbol{s_{mcp}}}^2} \tag{S21}$$

$$\theta'_{mcp} = \arcsin\frac{d_{mcp}}{|\boldsymbol{l_h}|} + \arccos\frac{|l_{b3}|^2+|l_h|^2-|l_{t3}|^2}{2|l_{b3}||l_h|}, \theta'_{mcp} \in [0,\pi] \tag{S22}$$

$$\theta_{mcp} = \theta'_{mcp} - \theta_{mcp0} \tag{S23}$$

Use $\boldsymbol{s_{ip}}$ and $\theta_{mcp}$ to calculate $\theta_{ip}$. $|\boldsymbol{s_{ip0}}|$ is the original length of the retractable linkage, and $|\boldsymbol{s_{ip}}|$is the length of the retractable linkage after compression.

$$|\boldsymbol{l_k}| = \sqrt{|\boldsymbol{l_{mcp}}|^2 + |\boldsymbol{l_{b5}}|^2 - 2\cos(\phi - \theta_{mcp})|\boldsymbol{l_{mcp}}|\,|\boldsymbol{l_{b5}}|} \tag{S24}$$

$$\theta'_{ip} = \arccos\frac{|\boldsymbol{l_{b4}}|^2 + |\boldsymbol{l_k}|^2 - |\boldsymbol{s_{ip}}|^2}{2|\boldsymbol{l_{b4}}||\boldsymbol{l_k}|}, \theta'_{ip} \in [0, \pi] \tag{S25}$$

$$\lambda = \arccos\frac{|\boldsymbol{l_{mcp}}|^2 + |\boldsymbol{l_k}|^2 - |\boldsymbol{l_{b5}}|^2}{2|\boldsymbol{l_{mcp}}||\boldsymbol{l_k}|}, \lambda \in [0, \pi] \tag{S26}$$

$$\theta_{ip} = \theta'_{ip} + \gamma - \lambda - \pi \tag{S27}$$

The serial kinematics of Torch-Hand were formulated using the DH convention. Each finger was modeled as an open serial chain rooted at the palm-fixed frame and terminating at the fingertip frame. The joint vectors of the four fingers and the thumb are defined in the main text, and the complete DH parameters are listed in Supplementary Table S3.

For the $\boldsymbol{i}$-th revolute joint, the homogeneous transformation from frame $\boldsymbol{i-1}$ to frame $\boldsymbol{i}$ is

$$^{i-1}\mathbf{T}_i(\theta_i) = \begin{bmatrix} \cos\theta_i & -\sin\theta_i\cos\alpha_i & \sin\theta_i\sin\alpha_i & a_i\cos\theta_i \\ \sin\theta_i & \cos\theta_i\cos\alpha_i & -\cos\theta_i\sin\alpha_i & a_i\sin\theta_i \\ 0 & \sin\alpha_i & \cos\alpha_i & d_1 \\ 0 & 0 & 0 & 1 \end{bmatrix}. \tag{S28}$$

Here, $\theta_i$ is the anatomical joint variable, and $a_i$, $d_i$ and $\alpha_i$ are the corresponding DH geometric parameters. This transformation describes how each anatomical joint contributes to the pose of the downstream phalanx.

For each of the four fingers, the fingertip pose relative to the palm-fixed frame is obtained by cascading the four serial joint transformations:

$$^{P}\mathbf{T}^{f}_{tip}(\mathbf{q}_f) = {}^{P}\mathbf{T}^{f}_{0}\left(\prod_{i=1}^{4} {}^{i-1}\mathbf{T}^{f}_{i}(\theta_{f,i})\right){}^{4}\mathbf{T}^{f}_{tip} \tag{S29}$$

Here, $P$ denotes the palm-fixed frame, $f$ denotes one of the four fingers, ${}^{P}_{1}\mathbf{T}^{f}_{0}$ defines the fixed transformation from the palm frame to the base frame of finger $f$, and ${}^{4}\mathbf{T}^{f}_{tip}$ defines the fixed transformation from the distal joint frame to the fingertip frame.

The thumb is modeled in the same way as a five-degree-of-freedom serial chain:

$$^{P}\mathbf{T}^{t}_{tip}(\mathbf{q}_t) = {}^{P}\mathbf{T}^{t}_{0}\left(\prod_{i=1}^{5} {}^{i-1}\mathbf{T}^{t}_{i}(\theta_{t,i})\right){}^{5}\mathbf{T}^{t}_{tip} \tag{S30}$$

Here, $t$ denotes the thumb. The five transformations correspond sequentially to CMC (Rot), CMC (Ab/Ad), CMC (F/E), MCP (F/E) and IP (F/E). The first three transformations define the orthogonal CMC motion axes that support thumb opposition.

The fingertip pose can be extracted from the homogeneous transformation matrix as

$$^{P}\mathbf{T}_{tip} = \begin{bmatrix} {}^{P}\mathbf{R}_{tip} & {}^{P}\mathbf{p}_{tip} \\ \mathbf{0}_{1\times3} & 1 \end{bmatrix}, \tag{S31}$$

where ${}^{P}\mathbf{R}_{tip}$ is the fingertip orientation matrix and ${}^{P}\mathbf{p}_{tip}$ is the fingertip position vector expressed in the palm-fixed frame. Therefore, the fingertip position used for workspace characterization is

$$ {}^{P}\mathbf{p}_{tip} = [x_{tip} \quad y_{tip} \quad z_{tip}]^T. \tag{S32} $$

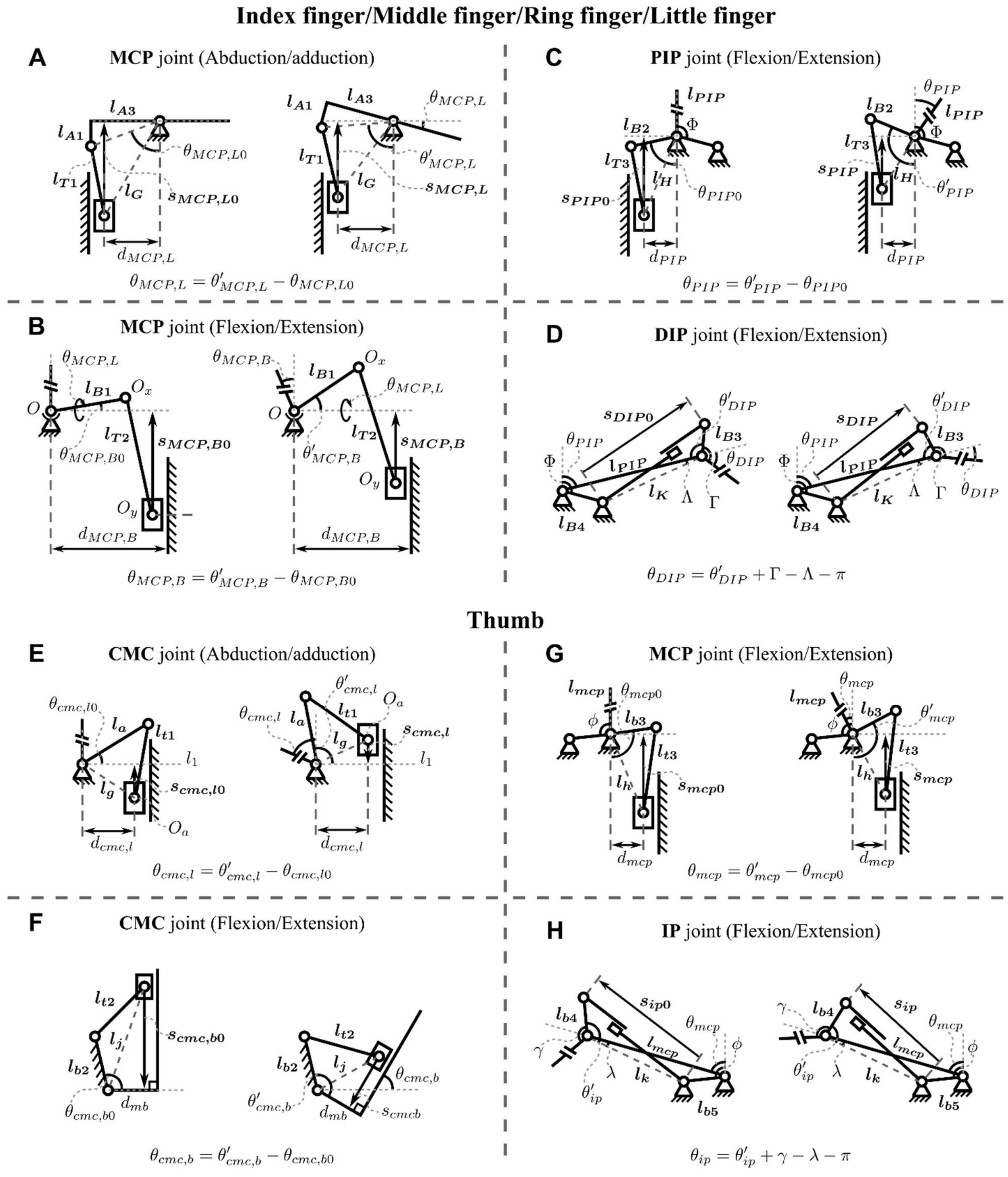

**Fig. S1. Geometric derivation of the slider-to-joint mapping in the slider–joint transmission.** (A to D) slider–crank linkage of the four fingers, including MCP (Ab/Ad), MCP (F/E), PIP (F/E), and DIP (F/E) joint. (E to H) slider–crank linkage of the thumb, including CMC (Ab/Ad), CMC (F/E), MCP (F/E), and IP (F/E) joint.

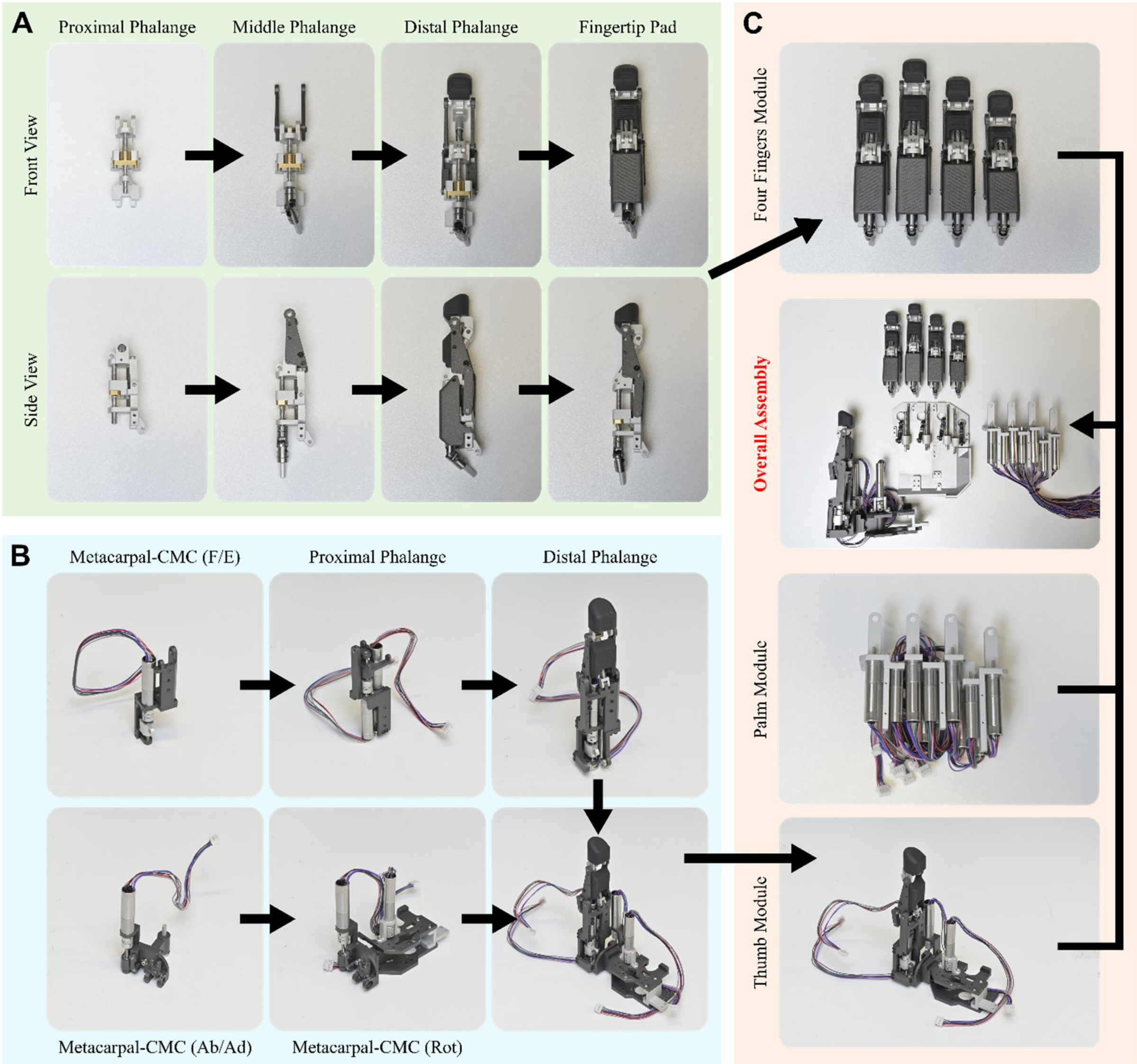

**Fig. S2. Modular palm–finger assembly workflow of Torch-Hand.** (A) Assembly workflow of the four-finger modules. (B) Assembly workflow of the thumb module. (C) Palm–finger integration.

.

**Table S1. Comparison between the existing humanoid dexterous hands and Torch-Hand.**

| Hand | Actuation paradigm | Joint | DOF | NOA | Load (kg) | Force (N) | Finnger size (mm) | Hand size (mm) | Remark |
|---|---|---|---|---|---|---|---|---|---|
| Torch-Hand | Linkage | 21 | 21 | 16 | >40 | >60 | 137.5 | 235 | / |
| LEAP Hand | Joint motor | 16 | 16 | 16 | 2 | 19.5 | 139.3 | 231.7 | / |
| LEAP Hand V2 | Joint motor | 17 | 21 | 17 | \ | \ | \ | \ | Force performance similar to LEAP Hand |
| Krysalis Hand | Joint motor | 20 | 18 | 18 | 4.53 | 10 | >120 | 240.5 | 2 DOF/Joint/Actuator in wrist |
| KITECH-Hand | Joint motor | 16 | 16 | 16 | 1.5 | / | / | / | / |
| DLR-HIT Hand | Gear | 13 | 13 | 13 | / | 7 | 135 | 234.2 | / |
| AVS hand | Gear | 8 | 8 | 8 | >3 | 30 | 120 | / | / |
| Multisensory Hand | Gear | 4 | 3 | 3 | 1.5 | 10 | 194 | / | One finger |
| ADAPT Hand | Tendon | 20 | 12 | 12 | / | ≈2.5 | >90 | >280 | / |
| F-TAC Hand | Tendon | 15 | 15 | 5 | 2.5 | 10.3 | >100 | 194 | / |
| BRSF Hand | Tendon | 12 | 8 | 8 | 1.75 | >2.8 | 105 | / | Biomimetic rigid-soft finger hand |
| VMS Hand | Tendon | 20 | 18 | 13 | / | / | 160 | 229 | / |
| CasiaHand | Tendon | 15 | 15 | 7 | >5 | 16.43 | 105 | 245 | / |
| BPI SoftHand | Tendon | 15 | 15 | 1 | >1.98 | 5.5 | 115 | 200 | / |
| OIS Hand | pneumatic | ∞ | ∞ | 5 | / | / | / | / | Optoelectronically Innervated Soft Hand |
| FHPA Hand | pneumatic | ∞ | ∞ | 5 | / | >10 | >115 | / | / |
| 3D-BJ Hand | pneumatic | 15 | 15 | 5 | >5.8 | 27.65 | 140 | / | 3D-Printed Buckling Joints hand |
| SRHB Hand | pneumatic | 15 | 15 | 10 | 1.86 | 7.5 | / | 205 | Soft-rigid hybrid bionic Hand |
| HBSF Hand | pneumatic | ∞ | ∞ | 5 | >0.541 | 2.44 | 92 | 197 | Grasp force in S11situation |
| ILDA Hand | Linkage | 20 | 15 | 15 | 18 | 34 | 101 | 261 | / |
| MCR-Hand III | Linkage | 21 | 21 | 16 | 2.5 | 7 | 108.31 | 205 | / |
| Prosthetic Hand | Linkage | 12 | 8 | 4 | / | 12 | 96 | / | Excluding the thumb |
| Slider-Cranks Finger | Linkage/Gear | 3 | 3 | 1 | / | >1.1 | 110 | / | Slider-Cranks finger (One finger) |
| SSSA-MyHand | Linkage/Gear | 10 | 4 | 3 | / | 14.6 | / | 200 | 31.4 N grasp force for the thumb |
| S-Finger Hand | Linkage/Gear | 2 | 1 | 1 | / | 24.2 | 83 | / | One finger |
| MSC Prosthetic Hand | Linkage/Tendon | 11 | 5 | 5 | / | 45 | / | / | / |
| SCD Hand | Linkage/Tendon | 15 | 5 | 5 | / | / | / | 215 | Simplified Configuration Design Hand |
| TSA Hand | Linkage/Tendon | 11 | 6 | 6 | / | 31.3 | 89 | / | |

The actuation paradigms reflect only the primary driving mechanisms of each hand; paradigms employed solely in one or two specialized joints are excluded from the corresponding hand's 'Actuation paradigm'. NOA represents the Number of Actuator

**Table S2. Joint and link definitions for elementary kinematic pairs in Torch-Hand.**

| Abbreviation | Full definition | Motion / Constraint | Functional role |
|---|---|---|---|
| R joint | Revolute joint | Single-axis rotation | Rotational joint |
| P joint | Prismatic joint | Linear translation | Sliding joint |
| H joint | Helical joint | Coupled rotation–translation | Screw transmission |
| S joint | Spherical joint | Three-axis rotation | Spatial rotation joint |
| U joint | Universal joint | Two-axis rotation | Spatial coupling joint |
| RR link | Revolute–Revolute link | Rotation–rotation connection | Binary linkage |
| SS link | Spherical–Spherical link | Spatial rotation connection | Spatial linkage |
| RPR link | Revolute–Prismatic–Revolute link | Rrtractable link | Length-adjustable link |

**Table S3. DH parameter table for the four fingers (Index finger/Middle finger/Ring finger/Little finger).**

| | $a_i$ | $\alpha_i$ | $d_i$ | $\theta_i$ |
|---|---|---|---|---|
| Joint 1 (MCP lateral motion joint) | 0 | $-90$ | 0 | $\theta_{MCP,L}$ |
| Joint 2 (MCP bend motion joint) | 0 | $-90$ | 0 | $\theta_{MCP,B}$ |
| Joint 3 (PIP bend motion joint) | $\lvert \boldsymbol{l_{MCP}} \rvert$ | 0 | 0 | $\theta_{PIP}$ |
| Joint 4 (DIP bend motion joint) | $\lvert \boldsymbol{l_{PIP}} \rvert$ | 0 | 0 | $\theta_{DIP}$ |

**Table S4: DH parameter table for the thumb.**

| | $a_i$ | $\alpha_i$ | $d_i$ | $\theta_i$ |
|---|---|---|---|---|
| Joint 1 (CMC rotation motion joint) | 0 | 0 | 0 | $\theta_{cmc,r}$ |
| Joint 2 (CMC lateral motion joint) | $\lvert \boldsymbol{l_{n1}} \rvert$ | $-90$ | $\lvert \boldsymbol{l'_r} \rvert$ | $\theta_{cmc,l}$ |
| Joint 3 (CMC bend motion joint) | $\lvert \boldsymbol{l'_{b1}} \rvert$ | 90 | $\lvert \boldsymbol{l_{n2}} \rvert$ | $\theta_{cmc,b}$ |
| Joint 4 (MCP bend motion joint) | $\lvert \boldsymbol{l_{cmc}} \rvert$ | 0 | 0 | $\theta_{mcp}$ |
| Joint 5 (DIP bend motion joint) | $\lvert \boldsymbol{l_{mcp}} \rvert$ | 0 | 0 | $\theta_{ip}$ |

**Table S5: Specifications of 3D-printed benchmark objects and corresponding real objects used in the 33-grasp taxonomy evaluation.**

| | Grasp taxonomy | 3D-printed object | Dimension (mm) | Real object | Mass (g) |
|---|---|---|---|---|---|
| 1 | Large Diameter | Large Cylinder | 180 | Milk Powder Can | 1090 |
| 2 | Small Diameter | Small Cylinder | 30 | Screwdriver | 285.5 |
| 3 | Medium Wrap | Middle Cylinder | 120 | Crisp Bucket | 159.8 |
| 4 | Adducted Thumb | Small Cylinder | 3 | Steel File | 170.2 |
| 5 | Light Tool | Thin cylinder | 12 | Glass Stirring Rod | 35.4 |
| 6 | Prismatic 4 Finger | Slender Cylinder | 15 | Spaghetti Strands | 141.1 |
| 7 | Prismatic 3 Finger | Slender Cylinder | 15 | Calligraphy Brush | 37.1 |
| 8 | Prismatic 2 Finger | Slender Cylinder | 15 | Plastic Straw | 2.6 |
| 9 | Palmar Pinch | Tiny Cylinder | 15 | Open-end Wrench | 139.6 |
| 10 | Power Disk | Middle Disk | 150 | Compact Disc | 14.0 |
| 11 | Power Sphere | Large Sphere | 100 | Gyroscopic Wrist Ball | 500.7 |
| 12 | Precision Disk | Large Disk | 165 | Packing Tape | 275.4 |
| 13 | Precision Sphere | Middle sphere | 75 | Orange | 150.2 |
| 14 | Tripod | Small Sphere | 50 | Golf Ball | 45.0 |
| 15 | Fixed Hook | Small Cylinder | 25 | Claw Hammer | 353.5 |
| 16 | Lateral | Small Thin Plate | 60 | Plastic Card | 6.6 |
| 17 | Index Finger Extension | Small Cylinder | 30 | Badminton Racket | 89.8 |
| 18 | Extension Type | Small Disk | 130 | Ceramic Dish | 310.9 |
| 19 | Distal Type | Articulated Object | Irregular | Scissors | 60.3 |
| 20 | Writing Tripod | Small Cylinder | 15 | Permanent Marker | 20.4 |
| 21 | Tripod Variation | Multi-component Object | 15 | Chopsticks | 22.5 |
| 22 | Parallel Extension | Large Thin Plate | 180 | Book | 132.6 |
| 23 | Adduction Grip | Slender Cylinder | 15 | Toothbrush | 14.5 |
| 24 | Tip Pinch | Miniscale Cylinder | 1.5 | Sewing Needle | 0.2 |
| 25 | Lateral Tripod | Small Sphere | 50 | Egg | 65.7 |
| 26 | Sphere 4 Finger | Large Sphere | 100 | Baseball | 130.1 |
| 27 | Quadpod | Middle Sphere | 75 | Peach | 116.8 |
| 28 | Sphere 3 Finger | Large Sphere | 100 | Tennis Ball | 60.2 |
| 29 | Stick | Slender Cylinder | 15 | Violin String | 70.7 |
| 30 | Palmar | Middle Disk | 150 | Flying Disc | 174.6 |
| 31 | Ring | Middle Cylinder | 120 | Glass Bottle | 550.4 |
| 32 | Ventral | Slender Cylinder | 15 | Utility Knife | 54.0 |
| 33 | Inferior Pincer | Tiny Sphere | 75 | Strawberry | 18.7 |

**Table S6: Data analysis of repeat positioning accuracy for Torch-Hand.**

| Finger | Joint | Average (N) | SD (N) | MR (N) | AD (N) | MD (N) |
|---|---|---|---|---|---|---|
| Thumb | CMC (Rot) | 0.0011 | 0.0139 | [-0.029,+0.031] | 0.0112 | +0.031 |
| | CMC (Ab/Ad) | 0.0013 | 0.0141 | [-0.025,+0.026] | 0.0117 | +0.026 |
| | CMC (F/E) | 0.0010 | 0.0149 | [-0.027,+0.026] | 0.0126 | -0.027 |
| | IP&MCP (F/E) | -0.0008 | 0.0119 | [-0.021,+0.023] | 0.0103 | +0.023 |
| Index | MCP (Ab/Ad) | -0.0002 | 0.0131 | [-0.024,+0.023] | 0.0111 | -0.024 |
| | MCP (F/E) | -0.0004 | 0.0133 | [-0.024,+0.025] | 0.0110 | +0.025 |
| | PIP&DIP (F/E) | -0.0012 | 0.0114 | [-0.023,+0.021] | 0.0096 | -0.023 |
| Middle | MCP (Ab/Ad) | 0.0004 | 0.0131 | [-0.025,+0.024] | 0.0109 | -0.025 |
| | MCP (F/E) | -0.0002 | 0.0153 | [-0.026,+0.025] | 0.0130 | -0.026 |
| | PIP&DIP (F/E) | -0.0003 | 0.0117 | [-0.021,+0.020] | 0.0100 | -0.021 |
| Ring | MCP (Ab/Ad) | -0.0014 | 0.0143 | [-0.023,+0.024] | 0.0125 | +0.024 |
| | MCP (F/E) | 0.0021 | 0.0138 | [-0.023,+0.025] | 0.0115 | +0.025 |
| | PIP&DIP (F/E) | 0.0001 | 0.0112 | [-0.020,+0.019] | 0.0095 | -0.020 |
| Little | (Ab/Ad) | 0.0008 | 0.0143 | [-0.022,+0.023] | 0.0127 | +0.023 |
| | MCP (F/E) | -0.0007 | 0.0127 | [-0.021,+0.024] | 0.0109 | +0.024 |
| | PIP&DIP (F/E) | -0.0005 | 0.0112 | [-0.017,+0.017] | 0.0100 | -0.017 |

SD represents standard deviation, MR represents measured range, AD represents average deviation, MD represents maximal deviation

**Table S7: Data analysis of joint-angle-dependent fingertip force output for Torch-Hand.**

| Finger | Joint | Average (N) | | | | | | | | | |
|---|---|---|---|---|---|---|---|---|---|---|---|
| | | 0° | 10° | 20° | 30° | 40° | 50° | 60° | 70° | 80° | 90° |
| Thumb | CMC (Rot) | 4.71 | 4.71 | 4.67 | 4.77 | 4.59 | 4.86 | 4.72 | 4.82 | 4.59 | / |
| | CMC (Ab/Ad) | 12.16 | 11.27 | 10.22 | 9.42 | 8.90 | 7.95 | 7.06 | 5.65 | 4.70 | 2.84 |
| | CMC (F/E) | 1.32 | 3.82 | 5.88 | 7.40 | 8.06 | 8.16 | 8.22 | 7.72 | 7.34 | 6.61 |
| | IP&MCP (F/E) | 33.59 | 33.97 | 32.00 | 30.08 | 27.83 | 27.24 | 24.86 | / | / | / |
| Index | MCP (Ab/Ad) | 41.74 | 42.2 | 40.96 | 39.50 | 36.36 | 33.81 | 31.07 | 26.70 | 22.03 | 15.28 |
| | MCP (F/E) | 33.95 | 32.68 | 30.93 | 27.94 | 25.84 | 22.09 | 19.86 | 15.13 | 11.96 | 9.18 |
| | PIP&DIP (F/E) | 60.36 | 61.72 | 56.43 | 56.32 | / | / | / | / | / | / |
| | MCP&PIP&DIP (F/E) | 7.18 | 7.04 | 6.63 | 5.45 | / | / | / | / | / | / |
| Middle | MCP (Ab/Ad) | 35.91 | 36.71 | 34.45 | 35.34 | 32.43 | 30.32 | 27.36 | 24.42 | 20.09 | 14.21 |
| | MCP (F/E) | 25.64 | 23.86 | 22.89 | 21.82 | 18.14 | 16.28 | 13.48 | 11.04 | 8.59 | 5.95 |
| | PIP&DIP (F/E) | 51.25 | 52.65 | 48.29 | 45.28 | / | / | / | / | / | / |
| | MCP&PIP&DIP (F/E) | 8.33 | 8.13 | 7.17 | 5.85 | / | / | / | / | / | / |
| Ring | MCP (Ab/Ad) | 41.32 | 42.86 | 41.39 | 40.12 | 36.07 | 34.98 | 31.59 | 27.38 | 22.26 | 16.08 |
| | MCP (F/E) | 34.47 | 32.30 | 30.24 | 27.26 | 25.71 | 23.17 | 19.69 | 15.40 | 12.31 | 8.83 |
| | PIP&DIP (F/E) | 60.30 | 57.61 | 56.87 | 55.14 | / | / | / | / | / | / |
| | MCP&PIP&DIP (F/E) | 9.55 | 8.76 | 7.96 | 6.60 | / | / | / | / | / | / |
| Little | (Ab/Ad) | 45.67 | 46.55 | 45.04 | 44.04 | 42.54 | 39.42 | 35.44 | 30.47 | 26.14 | 17.30 |

| | | | | | | | | | | |
|---|---|---|---|---|---|---|---|---|---|---|
| MCP (F/E) | 36.51 | 36.46 | 34.82 | 33.30 | 29.34 | 26.63 | 22.47 | 20.21 | 16.37 | 13.83 |
| PIP&DIP (F/E) | 63.68 | 64.69 | 61.41 | 58.17 | / | / | / | / | / | / |
| MCP&PIP&DIP (F/E) | 41.74 | 42.20 | 40.96 | 39.50 | 36.36 | 33.81 | 31.07 | 26.70 | 22.03 | 15.28 |